%% file: acl2021.tex
\documentclass[11pt,a4paper]{article}

\usepackage[hyphens]{url}
\usepackage[hyperref]{acl2021}
\usepackage{breakurl}
\hypersetup{breaklinks=true}
\usepackage{comment}
\usepackage{times}
\usepackage{latexsym}

\input{math_commands.tex}

\usepackage{url}
\usepackage{graphicx}
\usepackage{wrapfig}
\usepackage{multirow}
\usepackage{booktabs}

\usepackage{amsmath}
\usepackage{amsfonts}
\usepackage{amssymb}
\usepackage{dsfont}

\usepackage{microtype}
\usepackage{lipsum}

\aclfinalcopy 

\renewcommand{\sec}[1]{\hyperref[sec:#1]{Section~\ref*{sec:#1}}}
\newcommand{\app}[1]{\hyperref[app:#1]{Appendix~\ref*{app:#1}}}
\newcommand{\tab}[1]{\hyperref[tab:#1]{Table~\ref*{tab:#1}}}
\newcommand{\fig}[1]{\hyperref[fig:#1]{Figure~\ref*{fig:#1}}}
\newcommand{\eq}[1]{Eq.~\hyperref[eq:#1]{(\ref*{eq:#1})}}

\title{Measuring and Improving Model-Moderator Collaboration using Uncertainty Estimation}

\author{
Ian D.\ Kivlichan\thanks{~ Equal contribution; authors listed alphabetically.} \\
Jigsaw \\
\texttt{\small kivlichan@google.com}  \\
\And
Zi Lin$^*$\thanks{~ This work was done while Zi Lin was an AI resident at Google Research.} \\
Google Research \\
\texttt{\small lzi@google.com}  \\
\And
Jeremiah Liu$^*$ \\
Google Research \\
\texttt{\small jereliu@google.com}  \\
\And
Lucy Vasserman \\
Jigsaw \\
\texttt{\small lucyvasserman@google.com}
}

\begin{document}
\maketitle
\begin{abstract}
Content moderation is often performed by a collaboration between humans and machine learning models.
However, it is not well understood \emph{how} to design the collaborative process so as to maximize the combined moderator-model system performance.
This work presents a rigorous study of this problem, focusing on an approach that  incorporates model uncertainty into the collaborative process. 
First, we introduce principled metrics to describe the performance of the collaborative system under capacity constraints on the human moderator, quantifying how efficiently the combined system utilizes human decisions.
Using these metrics, we conduct a large benchmark study evaluating the performance of state-of-the-art uncertainty models under different collaborative review strategies. 
We find that an uncertainty-based strategy consistently outperforms the widely used strategy based on toxicity scores, and moreover that the choice of review strategy drastically changes the overall system performance.
Our results demonstrate the importance of rigorous metrics for understanding and developing effective moderator-model systems for content moderation, as well as the utility of uncertainty estimation in this domain.\footnote{Complete code including metric implementations and experiments is available at \url{http://github.com/google/uncertainty-baselines/tree/master/baselines/toxic_comments}.}
\end{abstract}

\input{sections/introduction}

\input{sections/related}

\input{sections/background}

\input{sections/task}

\input{sections/methods}
\input{sections/table_1}

\input{sections/experiments}

\input{sections/conclusion}

\input{sections/acknowledge}

\bibliographystyle{acl_natbib}
\bibliography{acl2021}

\clearpage
\appendix
\input{sections/appendix}

\end{document}

%% file: math_commands.tex
\usepackage{amsmath,amsfonts,bm}

\def\eqref#1{equation~\ref{#1}}

\def\1{\bm{1}}

\DeclareMathAlphabet{\mathsfit}{\encodingdefault}{\sfdefault}{m}{sl}
\SetMathAlphabet{\mathsfit}{bold}{\encodingdefault}{\sfdefault}{bx}{n}

\newcommand{\Dsc}{\mathcal{D}}

%% file: sections/introduction.tex
\section{Introduction}
\label{sec:intro}

Maintaining civil discussions online is a persistent challenge for online platforms.
Due to the sheer scale of user-generated text, modern content moderation systems often employ machine learning algorithms to automatically classify user comments based on their toxicity, with the goal of flagging a collection of likely policy-violating content for human experts to review \citep{etim2017times}.
However, modern deep learning models have been shown to suffer from reliability and robustness issues, especially in the face of the rich and complex sociolinguistic phenomena in real-world online conversations. Examples include possibly generating confidently wrong predictions based on spurious lexical features \citep{wang-culotta-2020-identifying}, or exhibiting undesired biases toward particular social subgroups \citep{46743}. 
This has raised questions about how current toxicity detection models will perform in realistic online environments, as well as the potential consequences for moderation systems \citep{rainie_future_2017}.

In this work, we study an approach to address these questions by incorporating model uncertainty into the collaborative model-moderator system's decision-making process. The intuition is that by using uncertainty as a signal for the likelihood of model error, we can improve the efficiency and performance of the collaborative moderation system by prioritizing 
the least confident examples from the model
for human review. 
Despite a plethora of uncertainty methods in the literature, there has been limited work studying their effectiveness in improving the performance of human-AI collaborative systems with respect to application-specific metrics and criteria \citep{awaysheh_review_2019, dusenberry_analyzing_2020, jesson_identifying_2020}. 
This is especially important for the content moderation task: real-world practice has unique challenges and constraints, including label imbalance, distributional shift, and limited resources of human experts; how these factors impact the collaborative system's effectiveness is not well understood. 

In this work, we lay the foundation for the study of the uncertainty-aware collaborative content moderation problem. We first (1) propose rigorous metrics
\textit{Oracle-Model Collaborative Accuracy} (OC-Acc) and \textit{AUC} (OC-AUC) 
to measure the performance of the overall collaborative system under capacity constraints on a simulated human moderator. We also propose \textit{Review Efficiency}, a intrinsic metric to measure a model's ability to improve the collaboration efficiency by selecting examples that need further review. Then, (2) we introduce a challenging data benchmark, \textit{\underline{Co}llaborative \underline{To}xicity \underline{Mo}deration in the Wil\underline{d}} (CoToMoD), for evaluating the effectiveness of a collaborative toxic comment moderation system. 
CoToMoD emulates the realistic train-deployment environment of a moderation system, in which the deployment environment contains richer linguistic phenomena and a more diverse range of topics than the training data, such that effective collaboration is crucial for good system performance \citep{amodei2016concrete}. Finally, (3) we present a large benchmark study to evaluate the performance of five classic and state-of-the-art uncertainty approaches on CoToMoD under two different moderation review approaches 
(based on the uncertainty score and on the toxicity score, respectively). 
We find that both the model's predictive and uncertainty quality contribute to the performance of the final system, and that the uncertainty-based review strategy 
outperforms the toxicity strategy across a variety of models and range of human review capacities. 

%% file: sections/related.tex
\section{Related Work}
\label{sec:related}




Our collaborative metrics draw on the idea of classification with a reject option, or learning with abstention \citep{JMLR:v9:bartlett08a,cortes2016learning,pmlr-v80-cortes18a,kompa2021second}.
In this classification scenario, the model has the option to reject an example instead of predicting its label.
The challenge in connecting learning with abstention to OC-Acc or OC-AUC is to account for how many examples have already been rejected.
Specifically, the difficulty is that the metrics we present are all dataset-level metrics, i.e.\ the ``reject'' option is not at the level of individual examples, but rather a set capacity over the entire dataset.
Moreover, this means OC-Acc and OC-AUC can be compared directly with traditional accuracy or AUC measures.
This difference in focus enables us to consider human time as the limiting resource in the overall model-moderator system's performance.

One key point for our work is that the best model (in isolation) may not yield the best performance in collaboration with a human \cite{bansal2021is}.
Our work demonstrates this for a case where the collaboration procedure is decided over the full dataset rather than per example: because of this, \citet{bansal2021is}'s expected team utility does not easily generalize to our setting.
In particular, the user chooses which classifier predictions to accept after receiving all of them rather than per example.

Robustness to distribution shift has been applied to toxicity classification in other works \citep{adragna2020fairness,koh2020wilds}, emphasizing the connection between fairness and robustness. Our work focuses on how these methods connect to the human review process, and how uncertainty can lead to better decision-making for a model collaborating with a human. 
Along these lines, \citet{dusenberry_analyzing_2020} analyzed how uncertainty affects optimal decisions in a medical context, though again at the level of individual examples rather than over the dataset.

%% file: sections/background.tex
\section{Background: Uncertainty Quantification for Deep Toxicity Classification}
\label{sec:background}

\noindent\textbf{Types of Uncertainty}~ Consider modeling a toxicity dataset $\Dsc=\{y_i, x_i\}_{i=1}^N$ using a deep classifier $f_W(x)$. Here the $x_i$ are example comments, $y_i \sim p^*(y|x_i)$ are toxicity labels drawn from a data generating process $p^*$ (e.g., the human annotation process), and $W$ are the parameters of the deep neural network. There are two distinct types of uncertainty in this modeling process: \textit{data uncertainty} and \textit{model uncertainty} \citep{sullivan2015introduction,liu_accurate_2019}. 
\textit{Data uncertainty} arises from the stochastic variability inherent in the data generating process $p^*$. For example, the toxicity label $y_i$ for a comment can vary between 0 and 1 depending on raters' different understandings of the comment or of the annotation guidelines.  On the other hand, \textit{model uncertainty} arises from the model's lack of knowledge about the world, commonly caused by insufficient coverage of the  training data. For example, at evaluation time, the toxicity classifier may encounter neologisms or misspellings that did not appear in the training data, making it more likely to make a mistake \citep{van-aken-etal-2018-challenges}. While the \textit{model uncertainty} can be reduced by training on more data, the \textit{data uncertainty} is inherent to the data generating process and is irreducible. 

\noindent\textbf{Estimating Uncertainty}~ A model that quantifies its uncertainty well should properly capture both the data and the model uncertainties.
To this end, a learned deep classifier $f_W(x)$ describes the \textit{data uncertainty} via its predictive probability, e.g.:
$$p(y|x, W) = \mathrm{sigmoid}(f_W(x)), $$
which is conditioned on the model parameter $W$, and is commonly learned by minimizing the Kullback-Leibler (KL) divergence between the model distribution $p(y|x, W)$ and the empirical distribution of the data (e.g.\ by minimizing the cross-entropy loss \citep{Goodfellow-et-al-2016}). On the other hand, a deep classifier can quantify \textit{model uncertainty} by using probabilistic methods to learn the posterior distribution of the model parameters:
$$W \sim p(W).$$
This distribution over $W$ leads to a distribution over the predictive probabilities $p(y|x, W)$. As a result, at inference time, the model can sample model weights $\{W_m\}_{m=1}^M$ from the posterior distribution $p(W)$, and then compute the posterior sample of predictive probabilities $\{p(y|x, W_m)\}_{m=1}^M$. This allows the model to express its model uncertainty through the variance of the posterior distribution $\mathrm{Var}\big( p(y|x, W) \big)$. \sec{method} surveys popular probabilistic deep learning methods.

In practice, it is convenient to compute a single uncertainty score capturing both types of uncertainty. To this end, we can first compute the marginalized predictive probability:
\begin{align*}
    p(y|x) = \int p(y|x, W) p(W) \,dW
\end{align*}
which captures both types of uncertainty by marginalizing the data uncertainty $p(y|x, W)$ over the model uncertainty $p(W)$. We can thus quantify the overall uncertainty of the model by computing the predictive variance of this binary distribution:
\begin{align}
    u_\text{unc}(x) = p(y|x) \times (1 - p(y|x)).
    \label{eq:pred_var}
\end{align}
\vspace{-12pt}

\noindent\textbf{Evaluating Uncertainty Quality}~
A common approach to evaluate a model's uncertainty quality is to measure its \textit{calibration} performance, i.e., whether the model's predictive uncertainty is indicative of the predictive error \citep{pmlr-v70-guo17a}. As we shall see in experiments, traditional calibration metrics like the Brier score \citep{NEURIPS2019_8558cb40}
do not correlate well with the model performance in collaborative prediction. One notable reason is that the collaborative systems use uncertainty as a ranking score (to identify possibly wrong predictions), while metrics like Brier score only measure the uncertainty's ranking performance indirectly.

\begin{figure}[ht]
\begin{center}
\begin{tabular}{|c|c|c|c|}
\cline{3-4}
\multicolumn{2}{c}{} & \multicolumn{2}{|c|}{\textbf{Uncertainty}} \\ \cline{3-4}
\multicolumn{2}{c|}{} & Uncertain  & Certain \\ \hline
\multirow{2}{*}{\textbf{Accuracy}} & Inaccurate & TP & FN \\ \cline{2-4}
& Accurate & FP  & TN \\ \hline
\end{tabular}
\end{center}
\caption{Confusion matrix for evaluating uncertainty calibration. We describe the correspondence in the text.}
\label{fig:calib_confusion_matrix}
\vspace{-8pt}
\end{figure}

This motivates us to consider \textit{Calibration AUC}, a new class 
of calibration metrics that focus on the uncertainty score $u_\text{unc}(x)$'s ranking performance. 
This metric evaluates uncertainty estimation by recasting it as a binary prediction problem, where the binary label is the model's prediction error $\mathbb{I}(f(x_i) \neq y_i)$, and the predictive score is the model uncertainty. This formulation leads to a confusion matrix as shown in \fig{calib_confusion_matrix} \citep{krishnan2020improving}.  Here, the four confusion matrix variables take on new meanings: (1) True Positive (TP) corresponds to the case where the prediction is inaccurate and the model is uncertain, (2) True Negative (TN) to the accurate and certain case, (3) False Negative (FN) to the inaccurate and certain case (i.e., over-confidence), and finally (4) False Positive (FP) to the accurate and uncertain case (i.e., under-confidence). Now, consider having the model predict its testing error using model uncertainty. The precision (TP/(TP+FP)) measures the fraction of inaccurate examples where the model is uncertain, recall (TP/(TP+FN)) measures the fraction of uncertain examples where the model is inaccurate, and the false positive rate (FP/(FP+TN)) measures the fraction of under-confident examples among the correct predictions. Thus, the model's calibration performance can be measured by the area under the precision-recall curve (\textbf{\textit{Calibration AUPRC}}) and under the receiver operating characteristic curve (\textbf{\textit{Calibration AUROC}}) for this problem. 
It is worth noting that the calibration AUPRC is closely related to the intrinsic metrics for the model's collaborative effectiveness: we discuss this in greater detail for the \textit{Review Efficiency} in \sec{metrics} and  \app{calib-auprc-discussion}).
This renders it especially suitable for evaluating model uncertainty in the context of collaborative content moderation.

%% file: sections/task.tex
\section{The Collaborative Content Moderation Task}
\label{sec:task}
Online content moderation is a \emph{collaborative} process, performed by humans working in conjunction with machine learning models. For example, the model can select a set of likely policy-violating posts for further review by human moderators. In this work, we consider a setting where a neural model interacts with an ``oracle'' human moderator with limited capacity in moderating online comments. 
Given a large number of examples $\{x_i\}_{i=1}^n$, the model first generates the predictive probability $p(y|x_i)$ and review score $u(x_i)$ for each example. Then, the model sends a pre-specified number of these examples to human moderators according to the rankings of the review score $u(x_i)$, and relies on its prediction $p(y|x_i)$ for the rest of the examples. In this work, we make the simplifying assumption that the human experts act like an oracle, correctly labeling all comments sent by the model. 

\subsection{Measuring the Performance of the Collaborative Moderation System}
\label{sec:metrics}

Machine learning systems for online content moderation are typically evaluated using metrics like accuracy or area under the receiver operating characteristic curve (AUROC). These metrics reflect the origins of these systems in classification problems, such as for detecting / classifying online abuse, harassment, or toxicity \cite{yin2009detection,Dinakar_Reichart_Lieberman_2011,Cheng_Danescu-Niculescu-Mizil_Leskovec_2015,10.1145/3038912.3052591}. However, 
they do not capture the model's ability to effectively collaborate with human moderators, or the performance of the resultant collaborative system.

New metrics, both extrinsic and intrinsic \cite{10.5555/1641396.1641403}, are one of the core contributions of this work. 
We introduce extrinsic metrics describing the performance of the overall model-moderator collaborative system (Oracle-Model Collaborative Accuracy and AUC, analogous to the classic accuracy and AUC), and an intrinsic metric focusing on the model's ability to effectively collaborate with human moderators (Review Efficiency), i.e., how well the model selects the examples in need of further review.

\paragraph{Extrinsic Metrics: Oracle-model Collaborative Accuracy and AUC~} 
To capture the collaborative interaction between human moderators and machine learning models, we first propose \textbf{\textit{Oracle-Model Collaborative Accuracy} (OC-Acc)}. 
OC-Acc measures the combined accuracy of this collaborative process, subject to a limited review capacity $\alpha$ for the human oracle (i.e., the oracle can process at most $\alpha \times 100\%$ of the total examples). Formally, given a dataset $D=\{(x_i, y_i)\}_{i=1}^n$, for a predictive model $f(x_i)$ generating a review score $u(x_i)$, the Oracle-Model Collaborative Accuracy for example $x_i$ is \\[-3pt]
\resizebox{\linewidth}{!}{
  \begin{minipage}{\linewidth}
    \begin{align*}
    \mbox{OC-Acc}(x_i | \alpha) &=
    \begin{cases}
    1 & \mbox{if } u(x_i) > q_{1-\alpha}  \\
    \mathbb{I}(f(x_i)=y_i ) & \mbox{otherwise}
    \end{cases},
    \end{align*}
  \end{minipage}}\\[8pt]
Thus, over the whole dataset, $\mbox{OC-Acc}(\alpha) = \frac{1}{n} \sum_{i=1}^n \mbox{OC-Acc}(x_i | \alpha)$.
Here $q_{1-\alpha}$ is the $(1-\alpha)^\text{th}$ quantile of the model's review scores $\{u(x_i)\}_{i=1}^n$ over the entire dataset.
OC-Acc thus describes the performance of a collaborative system which defers to a human oracle when the review score $u(x_i)$ is high, and relies on the model prediction otherwise, capturing the real-world usage and performance of the underlying model in a way that traditional metrics fail to.

However, as an accuracy-like metric, OC-Acc 
relies on a set threshold on the prediction score. This limits the metric's ability in describing model performance when compared to threshold-agnostic metrics like AUC. Moreover, OC-Acc can be sensitive to the intrinsic class imbalance in the toxicity datasets, appearing overly optimistic for model predictions that are biased toward negative class, similar to traditional accuracy metrics
\citep{DBLP:journals/corr/abs-1903-04561}. 
Therefore in practice, we prefer the AUC analogue of Oracle-Model Collaborative Accuracy, which we term the \textbf{\textit{Oracle-Model Collaborative AUC} (OC-AUC)}. 
OC-AUC measures the same collaborative process as the OC-Acc,
where the model sends the predictions with the top $\alpha \times 100\%$ of review scores. 
Then, similar to the standard AUC computation, OC-AUC sets up a collection of classifiers with varying predictive score thresholds, each of which has access to the oracle exactly as for OC-Acc  \cite{10.1145/1143844.1143874}. Each of these classifiers sends the same set of examples to the oracle (since the review score $u(x)$ is threshold-independent), and the oracle corrects model predictions when they are incorrect given the threshold. The OC-AUC---both OC-AUROC and OC-AUPRC---can then be calculated over this set of classifiers following the standard AUC algorithms \cite{10.1145/1143844.1143874}.

\paragraph{Intrinsic Metric: Review Efficiency~} The metrics so far measure the performance of the overall collaborative system, which combines both the model's predictive accuracy and the model's effectiveness in collaboration. To understand the source of the improvement, we also introduce \textit{\textbf{Review Efficiency}}, an intrinsic metric focusing solely on the model's effectiveness in collaboration.
Specifically, \textit{Review Efficiency} is the proportion of examples sent to the oracle for which the model prediction would otherwise have been incorrect. This can be thought of as the model's precision in selecting inaccurate examples for further review (TP/(TP+FP) in \fig{calib_confusion_matrix}). 

Note that the system's overall performance (measured by the oracle-model collaborative accuracy) can be rewritten as a weighted sum of the model's original predictive accuracy and the Review Efficiency (RE):
\begin{align}
&\mbox{OC-Acc}(\alpha)= \mbox{Acc} + 
\alpha \times \mbox{RE}(\alpha)
\label{eq:oca_decomp}
\end{align}
\noindent where $\mbox{RE}(\alpha)$ is the model's review efficiency 
among all the examples whose review score $u(x_i)$ are greater than  $q_{1-\alpha}$ (i.e., those sent to human moderators).
Thus, a model with better predictive performance and higher review efficiency yields better performance in the overall system. The benefits of review efficiency become more pronounced as the review fraction $\alpha$ increases.
We derive \eq{oca_decomp} in \app{oca-decomp-derivation}.

\input{sections/data}

%% file: sections/data.tex
\subsection{CoToMoD: An Evaluation Benchmark for Real-world Collaborative Moderation}
\label{sec:cotomod}

In a realistic industrial setting, toxicity detection models are often trained on a well-curated dataset with clean annotations, and then deployed to an environment that contains a more diverse range of sociolinguistic phenomena, and additionally exhibits systematic shifts in the lexical and topical distributions when compared to the training corpus. 

To this end, we introduce a challenging data benchmark,  \textbf{\textit{\underline{Co}llaborative \underline{To}xicity \underline{Mo}deration in the Wil\underline{d}} (CoToMoD)}, to evaluate the performance of collaborative moderation systems in a realistic environment. CoToMoD consists of a set of \textit{train}, \textit{test}, and \textit{deployment} environments: the \textit{train} and \textit{test} environments consist 
of 200k comments from  Wikipedia discussion comments from 2004--2015 (the Wikipedia Talk Corpus \cite{10.1145/3038912.3052591}), and the \textit{deployment} environment 
consists of one million public comments appeared on approximately 50 English-language news sites across the world from 2015--2017 (the CivilComments dataset \cite{DBLP:journals/corr/abs-1903-04561}). This setup mirrors the real-world implementation of these methods, where robust performance under changing data is essential for proper deployment \citep{amodei2016concrete}. 

Notably, CoToMoD contains two data challenges often encountered in practice: (1) \textit{Distributional Shift}, i.e.\ the comments in the training and deployment environments cover different time periods and surround different topics of interest (Wikipedia pages vs.\ news articles). As the CivilComments corpus is much larger in size, it contains a considerable collection of long-tail phenomena (e.g., neologisms, obfuscation, etc.) that appear less frequently in the training data. (2) \textit{Class Imbalance}, i.e.\ the fact that most online content is not toxic 
\citep{10.1145/2998181.2998213,10.1145/3038912.3052591}.
This manifests in the datasets we use: 
roughly $2.5\%$ (50,350 / 1,999,514) of the examples in the CivilComments dataset, and $9.6\%$ (21,384 / 223,549) of the examples in Wikipedia Talk Corpus examples are toxic \citep{10.1145/3038912.3052591, DBLP:journals/corr/abs-1903-04561}. As we will show, failing to account for class imbalance can severely bias model predictions toward the majority (non-toxic) class, reducing the effectiveness of the collaborative system. 


%% file: sections/methods.tex
\section{Methods}
\label{sec:method}

\paragraph{Moderation Review Strategy} In measuring model-moderator collaborative performance, we consider two review strategies (i.e.\ using different review scores $u(x)$). 
First, we experiment with a common toxicity-based review strategy \cite{latin_america_moderates,NYT-case-study}. Specifically, the model sends comments for review in decreasing order of the predicted toxicity score (i.e., the predictive probability $p(y|x)$), equivalent to a review score $u_\text{tox}(x) = p(y|x)$. 
The second strategy is uncertainty-based: given $p(y|x)$, we use uncertainty as the review score, $u_\text{unc}(x)=p(y|x)(1-p(y|x))$ (recall \eq{pred_var}), 
so that the review score is maximized at $p(y|x)=0.5$, and decreases toward 0 as $p(x)$ approaches 0 or 1. Which strategy performs best depends on the toxicity distribution in the dataset and the available review capacity $\alpha$.

\paragraph{Uncertainty Models} 
We evaluate the performance of classic and the latest state-of-the-art probabilistic deep learning methods on the CoToMoD benchmark. 
We consider BERT$_{\texttt{base}}$ as the base model \citep{devlin-etal-2019-bert}, and select five methods based on their practical applicability for transformer models. Specifically, we consider (1) \textit{Deterministic} which computes the sigmoid probability $p(x)=\mathrm{sigmoid}( \mathrm{logit}(x) )$  of a vanilla BERT model  \citep{DBLP:conf/iclr/HendrycksG17}, 
(2) \textit{Monte Carlo Dropout} (\textit{MC Dropout}) which estimates uncertainty using the Monte Carlo average of $p(x)$ from 10 dropout samples \citep{pmlr-v48-gal16},  
(3) \textit{Deep Ensemble} which estimates uncertainty using the ensemble mean of $p(x)$ from 10 BERT models trained in parallel \citep{NIPS2017_9ef2ed4b}, (4) \textit{Spectral-normalized Neural Gaussian Process} (\textit{SNGP}), a recent state-of-the-art approach which improves a BERT model's uncertainty quality by transforming it into an approximate Gaussian process model \citep{NEURIPS2020_543e8374}, and (5) \textit{SNGP Ensemble}, which is the Deep Ensemble using SNGP as the base model. 

\paragraph{Learning Objective} To address class imbalance, we consider combining the uncertainty methods with \textit{Focal Loss} \citep{Lin_2017_ICCV}. Focal loss reshapes the loss function to down-weight ``easy'' negatives (i.e.\ non-toxic examples), thereby focusing training on a smaller set of more difficult examples, and empirically leading to improved predictive and uncertainty calibration performance on class-imbalanced datasets \citep{Lin_2017_ICCV, NEURIPS2020_aeb7b30e}.
We focus our attention on focal loss (rather than other approaches to class imbalance) because of how this impact on calibration interacts with our moderation review strategies.

%% file: sections/table_1.tex
\begin{table*}[ht!]
\begin{center}
\begin{sc}
\resizebox{0.99\textwidth}{!}{
\begin{tabular}{l l || c c c | c c || c c c | c c}
\toprule
& & \multicolumn{5}{c ||}{Testing Env (Wikipedia Talk)} & \multicolumn{5}{c }{Deployment Env (CivilComments)} \\
\toprule
& Model
& 
 AUROC $\uparrow$ 
& AUPRC $\uparrow$
& Acc.\ $\uparrow$
& Brier $\downarrow$
& Calib.\ AUPRC $\uparrow$ 
& AUROC $\uparrow$ 
& AUPRC $\uparrow$
& Acc.\ $\uparrow$
& Brier $\downarrow$
& Calib.\ AUPRC $\uparrow$ 
\\
\midrule
\multirow{5}{*}{\rotatebox[origin=c]{90}{XENT}} &
Deterministic   
& $0.9734$ & $ 0.8019$ & $ 0.9231$ & $ 0.0548$ & $0.4053$
& $0.7796$ & $0.6689$ & $0.9628$ & $0.0246$  & $0.3581$
\\
& SNGP          
& $0.9741$ & $ 0.8029$ & $0.9233$ & $ 0.0548$ & $0.4063$
& $0.7695$ & $0.6665$ & $0.9640$ & $0.0253$ & $0.3660$
\\
& MC Dropout    
& $0.9729$ & $ 0.8006$ & $ \textbf{0.9274}$ & $\textbf{0.0508}$ & $0.4020$
& $0.7806$ & $0.6727$ & $\textbf{0.9671}$ & $\mathbf{0.0241}$ & $\mathbf{0.3707}$
\\
& Deep Ensemble      
& $0.9738$ & $ 0.8074$ & $ 0.9231$ & $ 0.0544$ & $0.4045$
& $\textbf{0.7849}$ & $\textbf{0.6741}$ & $0.9625$ & $0.0242$  & $0.3484$
\\
& SNGP Ensemble 
& $\textbf{0.9741}$ & $\textbf{0.8045}$ & $ 0.9226$ & $ 0.0549$ & $\mathbf{0.4158}$
& $0.7749$ & $0.6719$ & $0.9633$ & $0.0248$ & $0.3655$
\\

\midrule
\multirow{5}{*}{\rotatebox[origin=c]{90}{Focal}} &
Deterministic   
& $0.9730$ & $0.8036$ & $0.9476$  & $0.0628$ & $0.3804$
& $0.8013$ & $0.6766$ & $\mathbf{0.9795}$ & $0.0377$ & $0.3018$
\\
& SNGP          
& $0.9736$ & $0.8076$ & $0.9455$ & $0.0388$  & $0.3885$
& $0.8003$ & $0.6820$ & $0.9784$ & $\textbf{0.0264}$ &  $0.3181$
\\
& MC Dropout    
& $0.9741$ & $0.8076$ & $0.9472$ & $0.0622$ & $\textbf{0.3890}$
& $0.8009$ & $0.6790$ & $0.9790$ & $0.0360$ & $0.3185$
\\
& Deep Ensemble      
& $0.9735$ & $0.8077$ & $\textbf{0.9479}$ & $0.0639$ & $0.3840$
& $\mathbf{0.8041}$ & $0.6814$ & $\mathbf{0.9795}$ & $0.0381$ & $0.3035$
\\
& SNGP Ensemble 
& $\mathbf{0.9742}$ & $\mathbf{0.8122}$ & $0.9467$  & $\mathbf{0.0379}$ & $0.3846$
& $0.8002$ & $\mathbf{0.6827}$ & $0.9790$ & $0.0266$ & $\textbf{0.3212}$
\\
\bottomrule
\end{tabular}
}
\end{sc}
\caption{Metrics for models evaluated on the testing environment (the Wikipedia Talk corpus, left) and deployment environment (the CivilComments corpus, right).
XENT (top) and Focal (bottom) indicate models trained with cross-entropy and focal losses, respectively. 
The best metric values for each loss function are shown in bold.
\label{tab:indomain_metrics}
}
\end{center}
\end{table*}

%% file: sections/experiments.tex
\section{Benchmark Experiments}
\label{sec:experiments}



We first examine the prediction and calibration performance of the uncertainty models alone (Section \ref{sec:pred_calib_results}).
For prediction, we compute the predictive accuracy (Acc) and the predictive AUC (both AUROC and AUPRC). For uncertainty, we compute the Brier score (i.e., the mean squared error between true labels and predictive probabilities, a standard uncertainty metric), and also the Calibration AUPRC (\sec{background}). 

We then evaluate the models' collaboration performance under both the uncertainty- and the toxicity-based review strategies (Section \ref{sec:collab_results}). For each model-strategy combination, we measure the model's  collaboration ability by computing Review Efficiency, and evaluate the performance of the overall collaborative system using Oracle-Model Collaborative AUROC (OC-AUROC). 
We evaluate all collaborative metrics over a range of human moderator review capacities, with their review fractions (i.e., fraction of total examples the model sends to the moderator for further review) ranging over $\{0.001, 0.005, 0.01, 0.02, 0.05, 0.1, 0.15, 0.20\}$.

Results on further uncertainty and collaboration metrics (Calibration AUROC, OC-Acc, OC-AUPRC, etc.) are in \app{collaborative_metrics_results}.

\begin{figure*}[ht]
\centering
\includegraphics[width=0.4\textwidth]{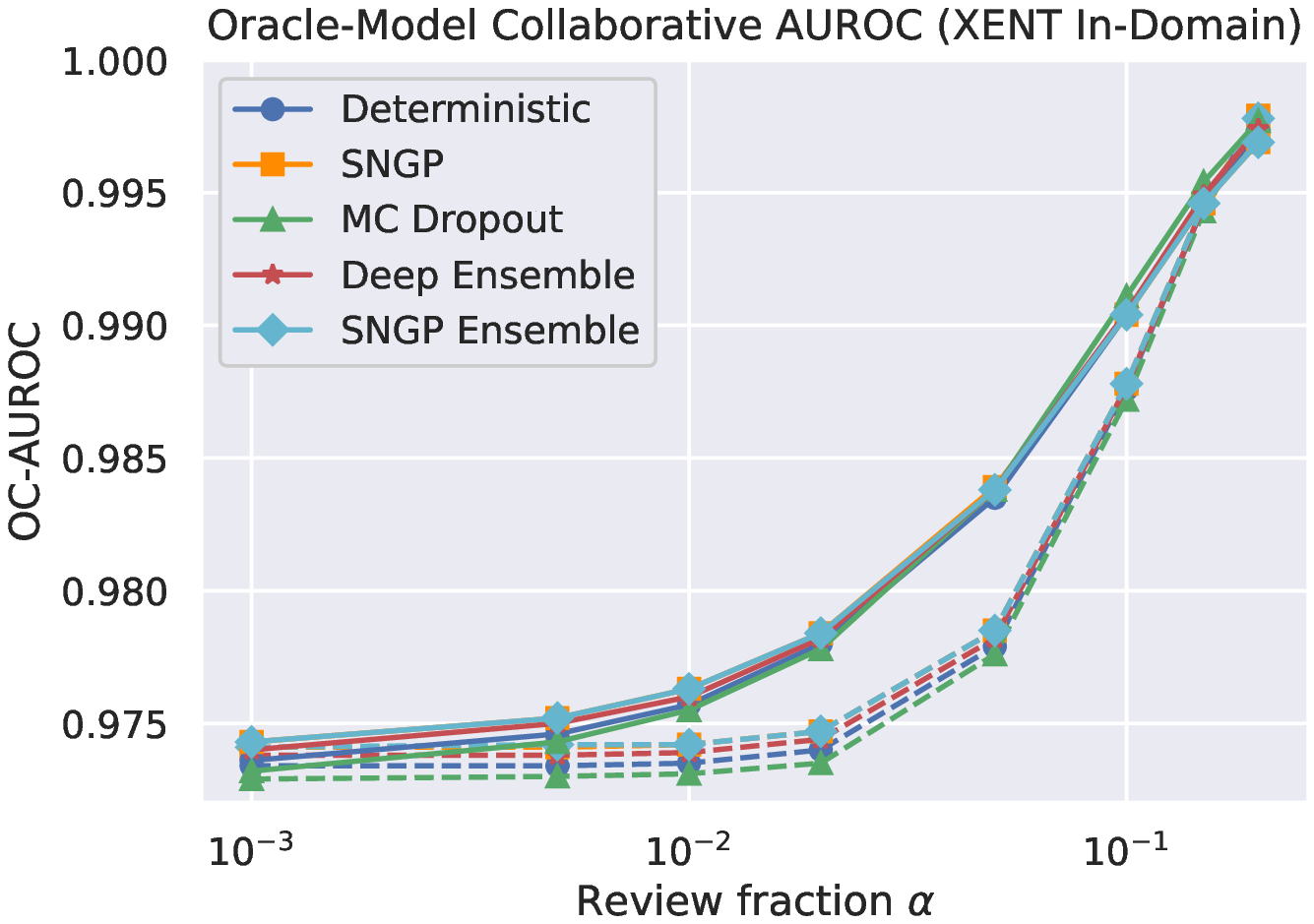}
\includegraphics[width=0.4\textwidth]{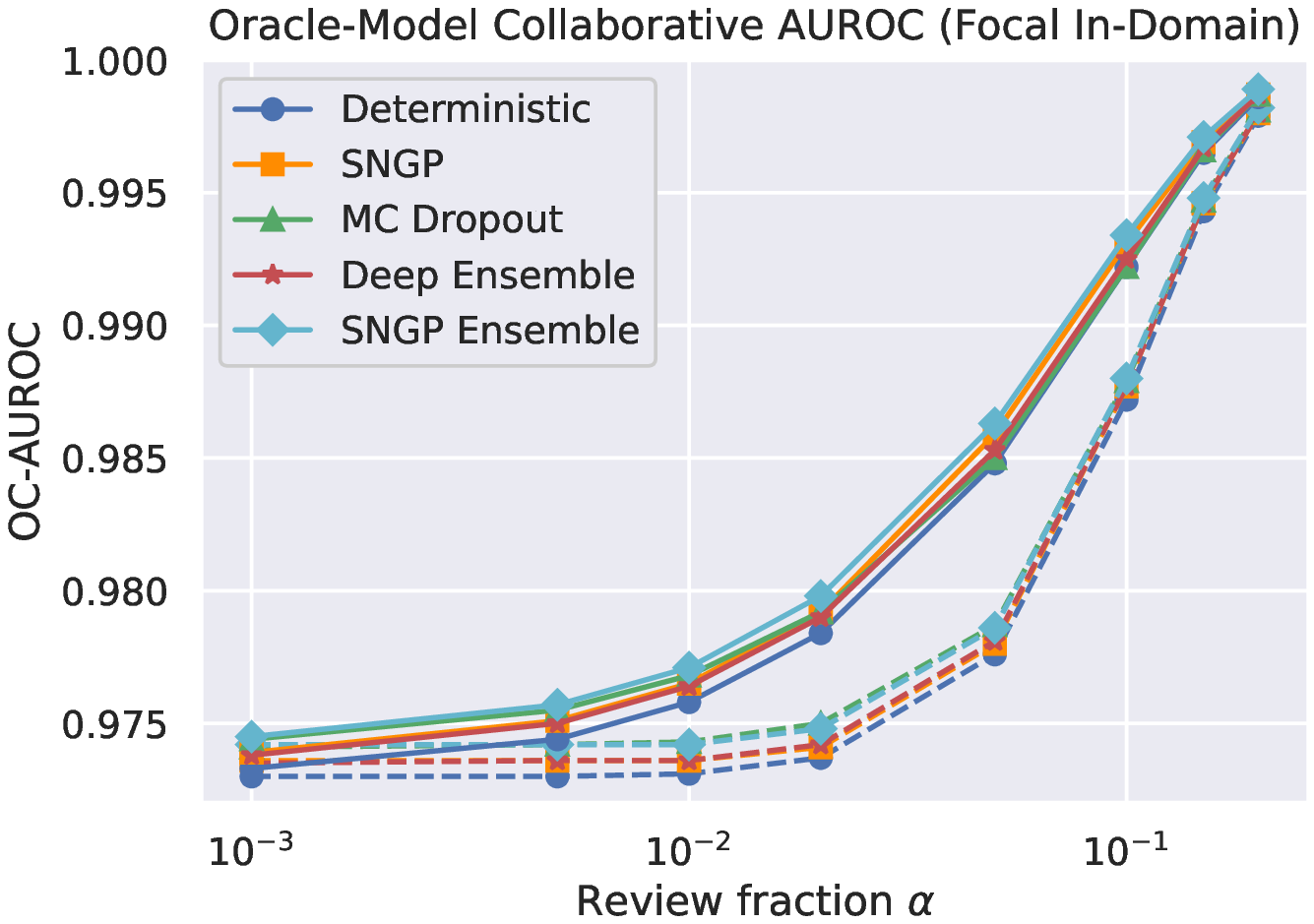}
\caption{Semilog plot of oracle-model collaborative AUROC as a function of review fraction (the proportion of comments the model can send for human/oracle review), trained with cross-entropy (XENT, left) or focal loss (right) and evaluated on the Wikipedia Talk corpus (i.e., the in-domain testing environment).
\textbf{Solid line:} uncertainty-based review strategy. \textbf{Dashed line:} toxicity-based review strategy. 
The best performing method is the SNGP Ensemble trained with focal loss and uses the uncertainty-based strategy.
\label{fig:ocauroc_ind_focal}
}
\vspace{-10pt}
\end{figure*}

\begin{figure*}[ht]
\centering
\includegraphics[width=0.4\textwidth]{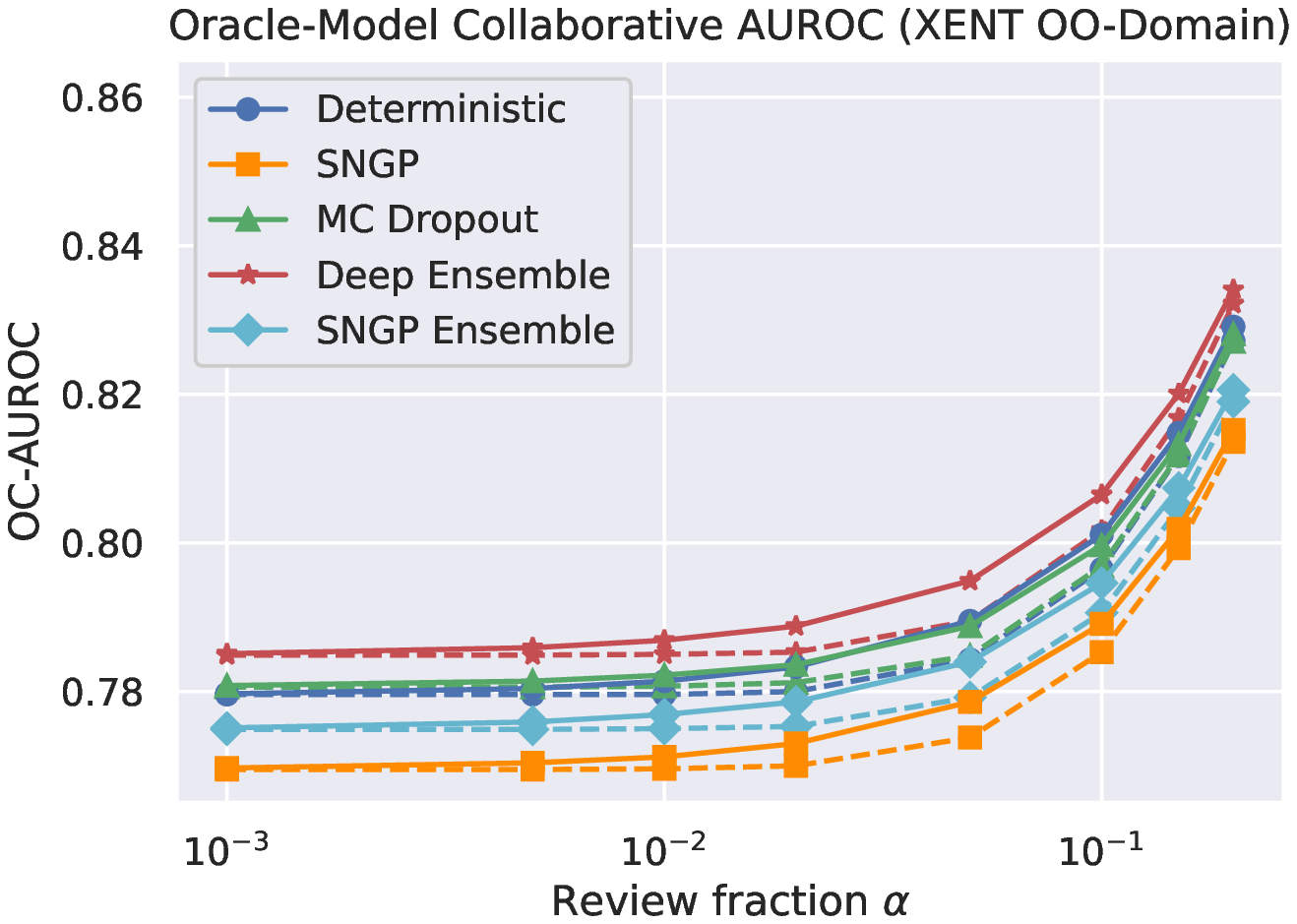}
\includegraphics[width=0.4\textwidth]{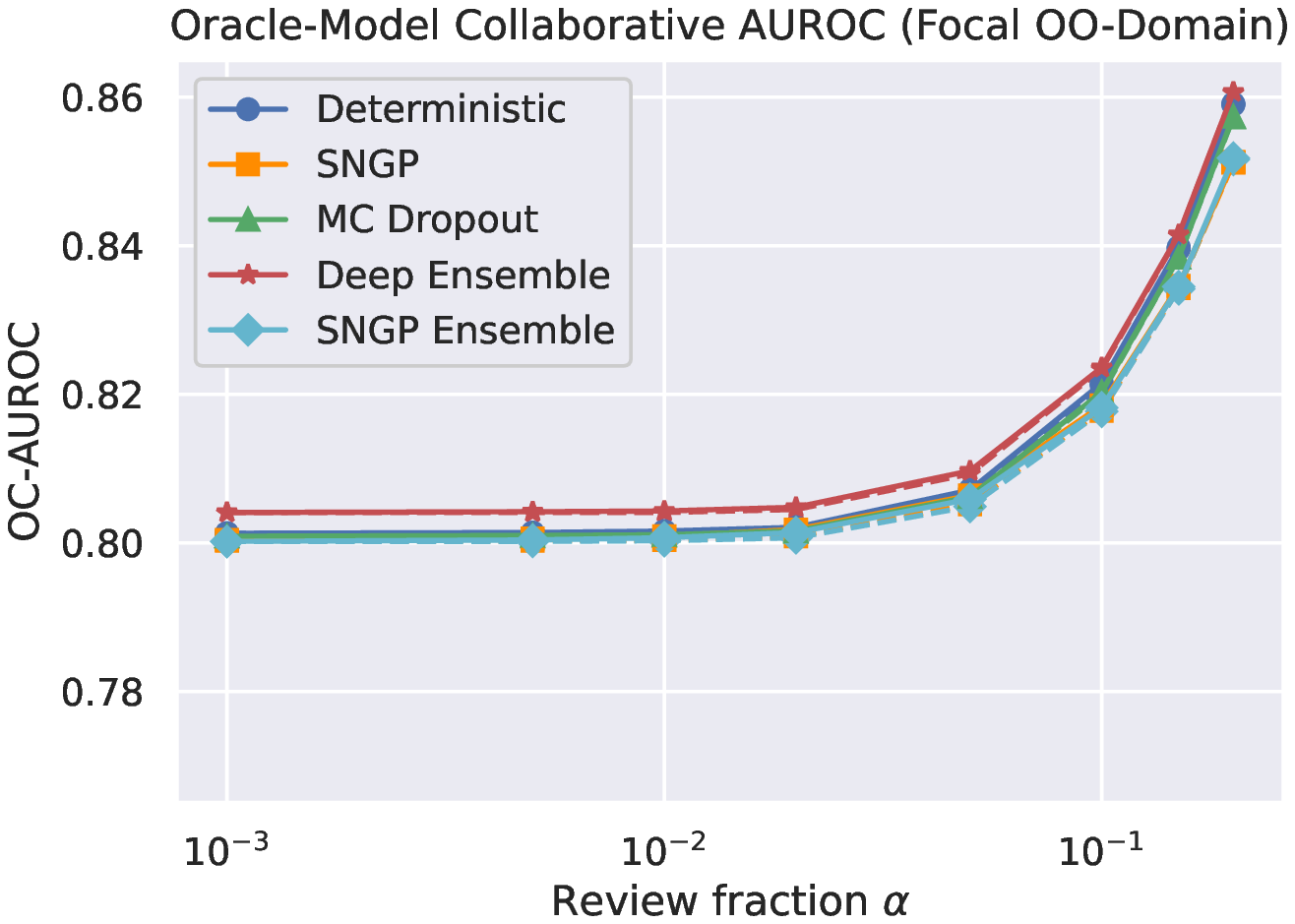}
\caption{Semilog plot of oracle-model collaborative AUROC as a function of review fraction, trained with cross-entropy (XENT, left) or focal loss (right) and evaluated on CivilComments corpus (i.e., the out-of-domain deployment environment). 
\textbf{Solid line:} uncertainty-based review strategy. \textbf{Dashed line:} toxicity-based review strategy. 
Training with focal rather than cross-entropy loss yields a large improvement. The best performing method is the Deep Ensemble trained with focal loss and uses the  uncertainty-based review strategy.
\label{fig:ocauroc_ood_focal}
}
\vspace{-10pt}
\end{figure*}

\begin{figure*}[ht]
\centering
\includegraphics[width=0.4\textwidth]{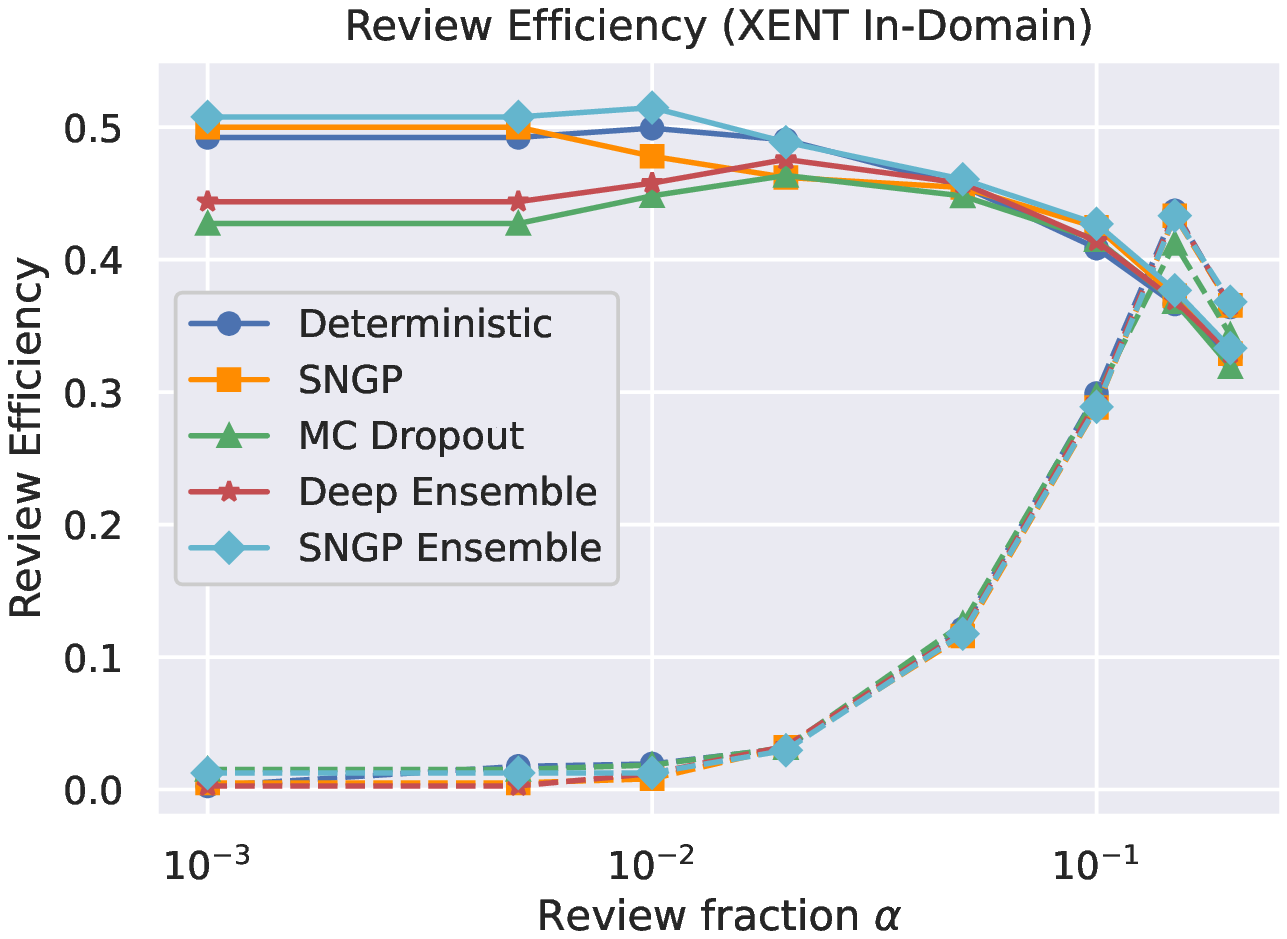}
\includegraphics[width=0.4\textwidth]{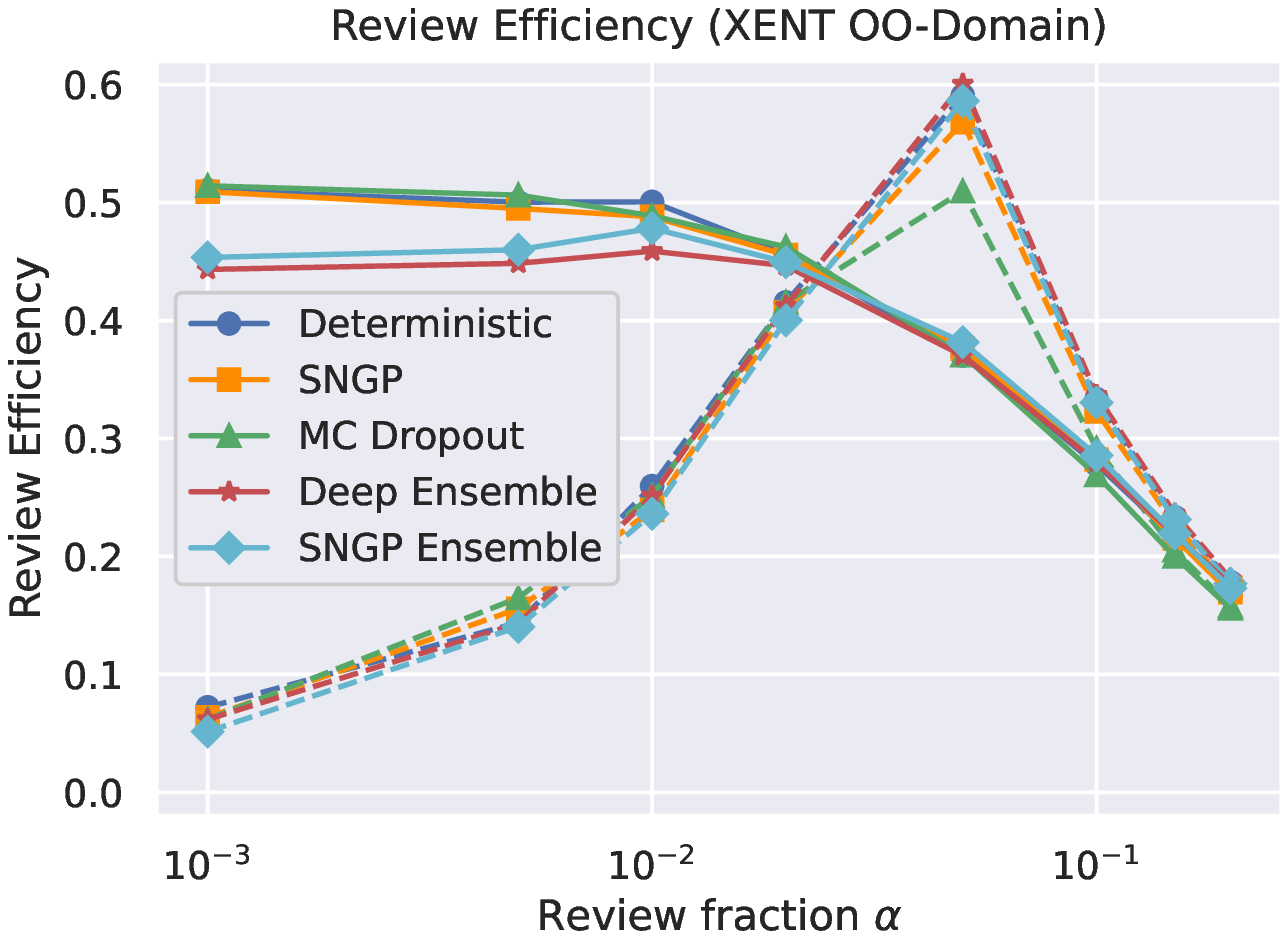}
\caption{
Semilog plot of review efficiency
as a function of review fraction, trained with cross-entropy and evaluated on the Wikipedia Talk corpus (i.e., the in-domain testing environment, left) and CivilComments (i.e., the out-of-domain deployment environment, right).
\textbf{Solid line:} uncertainty-based review strategy. \textbf{Dashed line:} toxicity-based review strategy. 
\label{fig:abstain_prec_ind_focal}
}
\vspace{-10pt}
\end{figure*}

\subsection{Prediction and Calibration}
\label{sec:pred_calib_results}

Table~\ref{tab:indomain_metrics} shows the performance of all uncertainty methods 
evaluated on the testing (the Wikipedia Talk corpus) and the deployment environments (the CivilComments corpus).

First, we compare the uncertainty methods based on the predictive and calibration AUC. As shown, for prediction, the ensemble models (both SNGP Ensemble and Deep Ensemble) provide the best performance, while the SNGP Ensemble and MC Dropout perform best for uncertainty calibration. Training with focal loss systematically improves the model prediction under class imbalance (improving the predictive AUC), while incurring a trade-off with the model's calibration quality (i.e.\ decreasing the calibration AUC).

Next, we turn to the model performance between the test and deployment environments. Across all methods, we observe a significant drop in predictive performance ($\sim\! 0.28$ for AUROC and $\sim\! 0.13$ for AUPRC), and a less pronounced, but still noticeable drop in uncertainty calibration ($\sim\! 0.05$ for Calibration AUPRC). Interestingly, focal loss seems to mitigate the drop in predictive performance, but also slightly exacerbates the drop in uncertainty calibration. 

Lastly, we observe a counter-intuitive improvement in the non-AUC metrics (i.e., accuracy and Brier score) in the out-of-domain deployment environment. This is likely due to their sensitivity to class imbalance 
(recall that toxic examples are slightly less rare in CivilComments). As a result, these classic metrics tend to favor model predictions biased toward 
the negative class, and therefore are less suitable for evaluating  model performance in the context of toxic comment moderation.


\subsection{Collaboration Performance}
\label{sec:collab_results}


\fig{ocauroc_ind_focal} and \ref{fig:ocauroc_ood_focal} show the Oracle-model Collaborative AUROC (OC-AUROC) of the overall collaborative system, and \fig{abstain_prec_ind_focal} shows the Review Efficiency of uncertainty models. Both the toxicity-based (dashed line) and uncertainty-based review strategies (solid line) are included. 

\paragraph{Effect of Review Strategy} For the AUC performance of the collaborative system, 
\textit{the uncertainty-based review strategy consistently outperforms the toxicity-based review strategy}. For example, in the in-domain environment (Wikipedia Talk corpus), using the uncertainty- rather than toxicity-based review strategy yields larger OC-AUROC improvements than any modeling change; this holds across all measured review fractions. We see a similar trend for OC-AUPRC (Appendix \fig{ocauprc_ind_focal}-\ref{fig:ocauprc_ood_focal}).

The trend in Review Efficiency (\fig{abstain_prec_ind_focal}) provides a more nuanced view to this picture. As shown, the efficiency of the toxicity-based strategy starts to improve as the review fraction increases, leading to a cross-over with the uncertainty-based strategy at high fractions. This is likely caused by the fact that in toxicity classification, 
the false positive rate exceeds the false negative rate.
Therefore sending a large number of positive predictions eventually leads the collaborative system to capture more errors, at the cost of a higher review load on human moderators. 
We notice that this transition occurs much earlier out-of-domain on CivilComments (\fig{abstain_prec_ind_focal} right). This highlights the impact of the toxicity distribution of the data on the best review strategy: 
because the proportion of toxic examples is much lower in CivilComments than in the Wikipedia Talk Corpus, the cross-over between the uncertainty and toxicity review strategies correspondingly occurs at lower review fractions.
Finally, it is important to note that this advantage in review efficiency does not directly translate to improvements for the overall system. For example, the OC-AUCs using the toxicity strategy are still lower than those with the uncertainty strategy even for high review fractions.

\paragraph{Effect of Modeling Approach} Recall that the performance of the overall collaborative system is the result of the model performance in both prediction and calibration, e.g.\ \eq{oca_decomp}. As a result, the model performance in Section \ref{sec:pred_calib_results} translates to performance on the collaborative metrics. 
For example, the ensemble methods (SNGP Ensemble and Deep Ensemble) consistently outperform on the OC-AUC metrics due to their high performance in predictive AUC and decent performance in calibration (\tab{indomain_metrics}). On the other hand, MC Dropout has good calibration performance but sub-optimal predictive AUC. As a result, it sometimes attains the best Review Efficiency (e.g., \fig{abstain_prec_ind_focal}, right), but never achieves the best overall OC-AUC. Finally, comparing between training objectives, the focal-loss-trained models tend to outperform their cross-entropy-trained counterparts in OC-AUC, due to the fact that focal loss tends to bring significant benefits to the predictive AUC (albeit at a small cost to the calibration performance).



%% file: sections/conclusion.tex
\section{Conclusion}


In this work, we presented the problem of collaborative content moderation, and introduced \textit{CoToMoD}, a challenging benchmark for evaluating the practical effectiveness of collaborative (model-moderator) content moderation systems. We proposed principled metrics to quantify how effectively a machine learning model and human (e.g.\ a moderator) can collaborate. These include \textit{Oracle-Model Collaborative Accuracy} (OC-Acc) and \textit{AUC} (OC-AUC), which measure analogues of the usual accuracy or AUC for interacting human-AI systems subject to limited human  review capacity. We also proposed \textit{Review Efficiency},
which quantifies how effectively a model utilizes human decisions.
These metrics are distinct from classic measures of predictive performance or uncertainty calibration, and enable us to evaluate the performance of the full collaborative system as a function of human attention, as well as to understand how efficiently the collaborative system utilizes human decision-making. 
Moreover, though we focused here on measuring the combined system's performance through metrics analogous to accuracy and AUC, it is trivial to extend these to other classic metrics like precision and recall.

Using these new metrics, we evaluated the performance of a variety of models on the collaborative content moderation task. We 
considered two canonical strategies for collaborative review: one based on the toxicity scores, and a new one using model uncertainty. 
We found that the uncertainty-based review strategy outperforms the toxicity strategy across a variety of models and range of human review capacities, yielding a $>\!\!30\%$ absolute increase in how efficiently the model uses human decisions and $\sim\!0.01$ and $\sim\!0.05$ absolute increases in the collaborative system's AUROC and AUPRC, respectively. 
This merits further study and consideration of this strategy's use in content moderation.
The interaction between the data distribution and best review strategy demonstrated by the crossover between the two strategies' performance out-of-domain) emphasizes the implicit trade-off between false positives and false negatives in the two review strategies: because toxicity is rare, prioritizing comments for review in order of toxicity reduces the false positive rate while potentially increasing the false negative rate.
By comparison, the uncertainty-based review strategy treats false positives and negatives more evenly.
Further study is needed to clarify this interaction.
Our work shows that the choice of review strategy drastically changes the collaborative system performance: 
evaluating and striving to optimize only the model yields much smaller improvements than changing the review strategy, and misses major opportunities to improve the overall system. 

Though the results presented in the current paper are encouraging, there remain important challenges for uncertainty modeling in the domain of toxic content moderation. 
In particular, dataset bias remains a significant issue: statistical correlation between the annotated toxicity labels and various surface-level cues may lead models to learn to overly rely on e.g.\ lexical or dialectal patterns \citep{zhou-etal-2021-challenges}. 
This could cause the model to produce high-confidence mispredictions for comments containing these cues (e.g., reclaimed words or counter-speech), resulting in a degradation in calibration performance in the deployment environment (cf.\ \tab{indomain_metrics}).  
Surprisingly, the standard debiasing techniques we experimented in this work (specifically, focal loss \citep{karimi-mahabadi-etal-2020-end}) only exacerbated this decline in calibration performance. This suggests that naively applying debiasing techniques may incur unexpected negative impacts on other aspects of the moderation system. 
Further research is needed into modeling approaches that can achieve robust performance both in prediction and in uncertainty calibration under data bias and distributional shift \citep{NEURIPS2020_eddc3427, utama-etal-2020-towards, DBLP:journals/corr/abs-2103-06922, yaghoobzadeh-etal-2021-increasing, bao2021predict, karimi-mahabadi-etal-2020-end}.

There exist several important directions for future work.
One key direction is to develop better review strategies than the ones discussed here: though the uncertainty-based strategy 
outperforms the toxicity-based one, there may be room for further improvement.
Furthermore, constraints on the moderation process may necessitate different review strategies: for example, if content can only be removed with moderator approval, we could experiment with a hybrid strategy which sends a mixture of high toxicity and high uncertainty content for human review.
A second direction is to study how these methods perform with real moderators: the experiments in this work are computational and there may exist further challenges in practice. For example, 
the difficulty of rating a comment can depend on the text itself in unexpected ways. 
Finally, a linked question is how to communicate uncertainty and different review strategies to moderators: simpler communicable strategies may be preferable to more complex ones with better theoretical performance.

%% file: sections/acknowledge.tex
\section*{Acknowledgements}

The authors would like to thank Jeffrey Sorensen for extensive feedback on the manuscript, and Nitesh Goyal, Aditya Gupta, Luheng He, Balaji Lakshminarayanan, Alyssa Lees, and Jie Ren for helpful comments and discussions.

%% file: sections/appendix.tex
\renewcommand{\textfraction}{0.05}

\section{Details on Metrics}
\label{app:metrics}

\subsection{Expected Calibration Error}
\label{app:ece-defn}

For completeness, we include a definition of the expected calibration error (ECE) \citep{10.5555/2888116.2888120} here. We use the ECE as a comparison for the uncertainty calibration performance alongside the Brier score in the tables in \app{collaborative_metrics_results}.

ECE can be computed by discretizes the probability range $[0, 1]$ into a set of $B$ bins, and computes the weighted average of the difference between confidence (the mean probability within each bin) and the accuracy (the fraction of predictions within each bin that are correct),
\begin{equation}
\text{ECE} = \sum_{b=1}^B \frac{n_b}{N} |\text{conf}(b) - \text{acc}(b)|,
\end{equation}
where $\text{acc}(b)$ and $\text{conf}(b)$ denote the accuracy and confidence for bin $b$, respectively, $n_b$ is the number of examples in bin $b$, and $N = \sum_b n_b$ is the total number of examples.

\subsection{Connection between Calibration AUPRC and Collaboration Metrics}
\label{app:calib-auprc-discussion}

As discussed in Section \ref{sec:background}, Calibration AUPRC is an especially suitable metric for measuring model uncertainty in the context of collaborative content moderation, due to its close connection with the intrinsic metrics for the model's collaboration effectiveness. 

Specifically, the \textit{Review Efficiency} metric (introduced in Section \ref{sec:metrics}) can be understood as the analog of \textbf{precision} for the calibration task. To see this, recall the four confusion matrix variables introduced in Figure \ref{fig:calib_confusion_matrix}: (1) True Positive (TP) corresponds to the case where the prediction is inaccurate and the model is uncertain, (2) True Negative (TN) to the accurate and certain case, (3) False Negative (FN) to the inaccurate and certain case (i.e., over-confidence), and finally (4) False Positive (FP) to the accurate and uncertain case (i.e., under-confidence).

Then, given a review capacity constraint $\alpha$, we see that
\begin{align*}
    \mbox{ReviewEfficiency}(\alpha) = \frac{TP_\alpha}{TP_\alpha + FP_\alpha},
\end{align*}
which measures the proportion of examples that were sent to human moderator that would otherwise be classified incorrectly. 

Similarly, we can also define the analog of \textbf{recall} for the calibration task, which we term \textit{Review Effectiveness}:
\begin{align*}
    \mbox{ReviewEffectiveness}(\alpha) = \frac{TP_\alpha}{TP_\alpha + FN_\alpha}.
\end{align*}
Review Effectiveness is also a valid intrinsic metric for the model's collaboration effectivess. It measures the proportion of incorrect model predictions that were successfully corrected using the review strategy. (We visualize model performance in Review Effectiveness in Section \ref{app:collaborative_metrics_results}.)

To this end, the calibration AUPRC can be understood as the area under the Review Efficiency v.s.\ Review Effectiveness curve, with the usual classification threshold replaced by the review capacity $\alpha$. Therefore, calibration AUPRC serves as a threshold-agnostic metric that captures the model's intrinsic performance in collaboration effectiveness.


\subsection{Further Discussion}


For the uncertainty-based review, an important question is whether classic uncertainty metrics like Brier score capture good model-moderator collaborative efficiency.
The SNGP Ensemble's good performance contrasts with its poorer Brier score (Table~\ref{tab:indomain_metrics}). 
By comparison, the calibration AUPRC successfully captures this good performance, and is highest for that model.
More generally, the low-review fraction review efficiency with cross-entropy is exactly captured by the calibration AUPRC (same ordering for the two measures).
This correspondence is not perfect: though the SNGP Ensemble with focal loss has the highest review efficiency overall, its calibration AUPRC is lower than the MC Dropout or SNGP models (models with next highest review efficiencies).
This may reflect the reshaping effect of focal loss on SNGP's calibration (explored in \app{reliability}).
Overall, calibration AUPRC much better captures the relationship between collaborative ability and calibration than do classic calibration metrics like Brier score (or ECE, see \app{collaborative_metrics_results}).
This is because classic calibration metrics are population-level averages, whereas calibration AUPRC measures the ranking of the predictions, and is thus more closely linked to the review order problem.

\section{Connecting Review Efficiency and Collaborative Accuracy}
\label{app:oca-decomp-derivation}
In this appendix, we derive \eq{oca_decomp} from the main paper, which connects the Review Efficiency and Oracle-Collaborative Accuracy.

Given a trained toxicity model, a review policy and a dataset, let us denote $r$ as the event that an example gets reviewed, and $c$ as the event that model prediction is correct. Now, assuming the model sends $\alpha \times 100\%$ of examples for human review, we have:
\begin{align*}
    \mbox{Acc} &= P(c), \qquad \alpha = P(r).
\end{align*}
Also, we can write:
\begin{align*}
    \mbox{RE}(\alpha) &= P(\neg c | r)
\end{align*}
i.e., review efficiency $\mbox{RE}(\alpha)$ is the percentage of incorrect predictions among reviewed examples. Finally:
\begin{align*}
    \mbox{OC-Acc}(\alpha) 
    &= P(c \cap \neg r) + P(c \cap r) + P(\neg c \cap r)
\end{align*}
i.e., an example is predicted correctly by the collaborative system if either the model prediction itself is accurate ($c \cap \neg r$), or it was sent for human review ($c \cap r$ or $\neg c \cap r$). 

The above expression of $\mbox{OC-Acc}$ leads to two different decompositions of the OC-Acc. First,
\begin{align*}
\mbox{OC-Acc}(\alpha) &= P(c \cap \neg r) + P(r) \\
&= P(c|\neg r)P(\neg r) + P(r) \\
&= \mbox{Acc}(1-\alpha) * (1-\alpha) + \alpha,
\end{align*}
where $\mbox{Acc}(1-\alpha)$ is the accuracy among the $(1-\alpha) \times 100\%$ examples that are not sent to human for review. 

\noindent Alternatively, we can write
\begin{align*}
\mbox{OC-Acc}(\alpha) 
&= P(c) + P(\neg c \cap r) \\
&= P(c) + P(\neg c | r) P(r) \\
&= \mbox{Acc} + \mbox{RE}(\alpha) * \alpha,
\end{align*}
which coincides with the expression in \eq{oca_decomp}.

\section{Reliability Diagrams for Deterministic and SNGP models}
\label{app:reliability}

\begin{figure*}[ht]
\centering
\includegraphics[width=\textwidth]{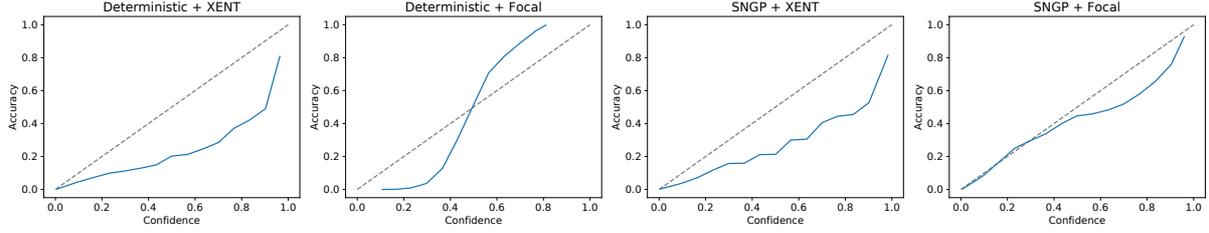}
\caption{ In-domain reliability diagrams for deterministic models and SNGP models with cross-entropy (XENT) and focal cross-entropy.
\label{fig:ece-curve-ind}
}
\end{figure*}

\begin{figure*}[ht]
\centering
\includegraphics[width=\textwidth]{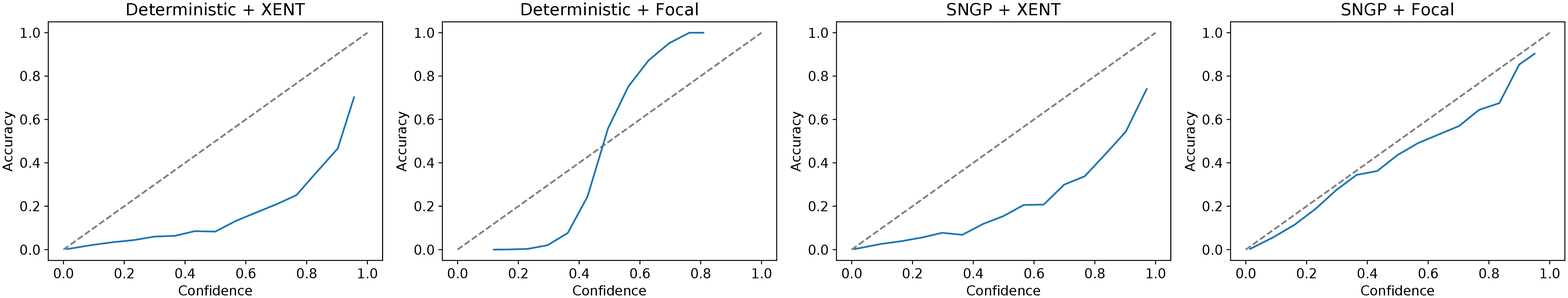}
\caption{Reliability diagrams for deterministic models and SNGP models with cross-entropy (XENT) and focal cross-entropy on the CivilComments dataset.
\label{fig:ece-curve-ood}
}
\end{figure*}

We study the effect of focal loss on calibration quality for SNGP in further detail. We plot the reliability diagrams for the deterministic and SNGP models trained with cross-entropy and focal cross-entropy. \fig{ece-curve-ind} shows the reliability diagrams in-domain and \fig{ece-curve-ood} shows them out-of-domain. 
We see that focal loss fundamentally changes the models' uncertainty behavior, systematically shifting the uncertainty curves from overconfidence (the lower right, below the diagonal) and toward the calibration line (the diagonal).  
However, the exact pattern of change is model dependent. We find that the deterministic model with focal loss is over-confident for predictions under $0.5$, and under-confident above $0.5$, while the SNGP models are still over-confident, although to a lesser degree compared to using cross-entropy loss.

\begin{table*}[ht!]
\begin{center}
\begin{sc}
\resizebox{0.8\textwidth}{!}{
\begin{tabular}{l l | c c c | c c | c c}
\toprule
& Model (Test)
& 
 AUROC $\uparrow$ 
& AUPRC $\uparrow$
& Acc.\ $\uparrow$
& ECE $\downarrow$
& Brier $\downarrow$
& Calib.\ AUROC $\uparrow$
& Calib.\ AUPRC  $\uparrow$
\\
\midrule
\multirow{5}{*}{\rotatebox[origin=c]{90}{XENT}} &
Deterministic   & $0.9734$ & $ 0.8019$ & $ 0.9231$ & $0.0245$ & $ 0.0548$ & $0.9230$ & $0.4053$ \\
& SNGP          & $0.9741$ & $ 0.8029$ & $ 0.9233$ & $0.0280$ & $ 0.0548$ & $0.9238$ & $0.4063$ \\
& MC Dropout    & $0.9729$ & $ 0.8006$ & $ \mathbf{0.9274}$ & $\mathbf{0.0198}$ & $ \mathbf{0.0508}$ & $\textbf{0.9282}$ & $0.4020$ \\
& Deep Ensemble      & $0.9738$ & $ \mathbf{0.8074}$ & $ 0.9231$ & $0.0235$ & $ 0.0544$ & $0.9245$ & $0.4045$ \\
& SNGP Ensemble & $\mathbf{0.9741}$ & $ 0.8045$ & $ 0.9226$ & $0.0281$ & $ 0.0549$ & $0.9249$ & $\mathbf{0.4158}$ \\

\midrule
\multirow{5}{*}{\rotatebox[origin=c]{90}{Focal}} &
Deterministic   & $0.9730$ & $0.8036$ & $0.9476$ & $0.1486$ & $0.0628$ & $0.9405$ & $0.3804$ \\
& SNGP          & $0.9736$ & $0.8076$ & $0.9455$ & $0.0076$ & $0.0388$ & $0.9385$ & $0.3885$ \\
& MC Dropout    & $0.9741$ & $0.8076$ & $0.9472$ & $0.1442$ & $0.0622$ & $\mathbf{0.9425}$ & $\textbf{0.3890}$ \\
& Deep Ensemble      & $0.9735$ & $0.8077$ & $\textbf{0.9479}$ & $0.1536$ & $0.0639$ & $0.9418$ & $0.3840$ \\
& SNGP Ensemble & $\mathbf{0.9742}$ & $\mathbf{0.8122}$ & $0.9467$ & $\mathbf{0.0075}$ & $\mathbf{0.0379}$ & $0.9400$ & $0.3846$ \\
\bottomrule
\end{tabular}
}
\end{sc}
\caption{Metrics for models on the Wikipedia Talk corpus (in-domain testing environment), all numbers are averaged over 10 model runs. 
XENT and Focal indicate models trained with the cross-entropy and focal losses, respectively. 
The best metric values for each loss function are shown in bold.
\label{tab:indomain_metrics_sep}
}
\end{center}
\end{table*}

\begin{table*}[t]
\begin{center}
\begin{sc}
\resizebox{0.8\textwidth}{!}{
\begin{tabular}{l l | c c c | c c | c c}
\toprule
& Model (Deployment)
& 
 AUROC $\uparrow$ 
& AUPRC $\uparrow$
& Acc.\ $\uparrow$
& ECE $\downarrow$
& Brier $\downarrow$
& Calib.\ AUROC $\uparrow$
& Calib.\ AUPRC $\uparrow$
\\
\midrule
\multirow{5}{*}{\rotatebox[origin=c]{90}{XENT}} &
Deterministic   & $0.7796$ & $0.6689$ & $0.9628$ & $0.0128$ & $0.0246$ & $0.9412$ & $0.3581$ \\
& SNGP          & $0.7695$ & $0.6665$ & $0.9640$ & $\textbf{0.0070}$ & $0.0253$ & $0.9457$ & $0.3660$ \\
& MC Dropout    & $0.7806$ & $0.6727$ & $\mathbf{0.9671}$ & $0.0136$ & $\mathbf{0.0241}$ & $\mathbf{0.9502}$ & $\mathbf{0.3707}$ \\
& Deep Ensemble      & $\mathbf{0.7849}$ & $\mathbf{0.6741}$ & $0.9625$ & $0.0141$ & $0.0242$ & $0.9420$ & $0.3484$ \\
& SNGP Ensemble & $0.7749$ & $0.6719$ & $0.9633$ & $0.0076$ & $0.0248$ & $0.9463$ & $0.3655$ \\

\midrule
\multirow{5}{*}{\rotatebox[origin=c]{90}{Focal}} &
Deterministic   & $0.8013$ & $0.6766$ & $\mathbf{0.9795}$ & $0.1973$ & $0.0377$ & $0.9444$ & $0.3018$ \\
& SNGP          & $0.8003$ & $0.6820$ & $0.9784$ & $0.0182$ & $\mathbf{0.0264}$ & $0.9465$ & $0.3181$ \\
& MC Dropout    & $0.8009$ & $0.6790$ & $0.9790$ & $0.1896$ & $0.0360$ & $\textbf{0.9481}$ & $0.3185$ \\
& Deep Ensemble      & $\mathbf{0.8041}$ & $0.6814$ & $\mathbf{0.9795}$ & $0.1998$ & $0.0381$ & $0.9461$ & $0.3035$ \\
& SNGP Ensemble & $0.8002$ & $\mathbf{0.6827}$ & $0.9790$ & $\mathbf{0.0176}$ & $0.0266$ & $\textbf{0.9481}$ & $\textbf{0.3212}$ \\
\bottomrule
\end{tabular}
}
\end{sc}
\caption{Metrics for models on the CivilComments corpus (out-of-domain deployment environment), all numbers are averaged over 10 model runs. 
XENT and Focal indicate models trained with the cross-entropy and focal losses, respectively. 
The best metric values for each loss function are shown in bold.
}
\label{tab:oodomain_metrics_sep}
\end{center}
\vspace{-0.75em}
\end{table*}

\section{Complete metric results}
\label{app:collaborative_metrics_results}

\begin{figure*}[ht]
\centering
\includegraphics[width=0.4\textwidth]{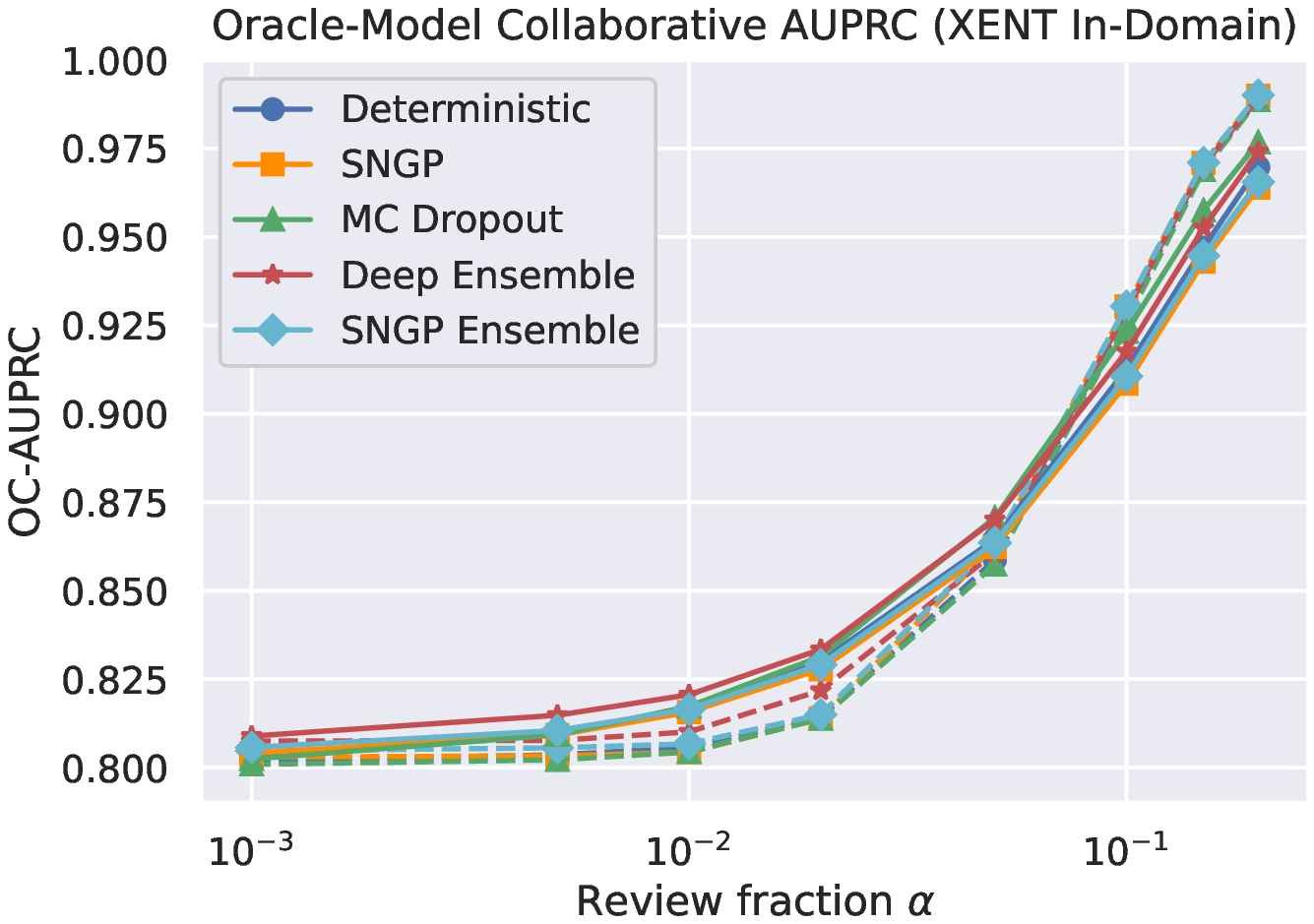}
\includegraphics[width=0.4\textwidth]{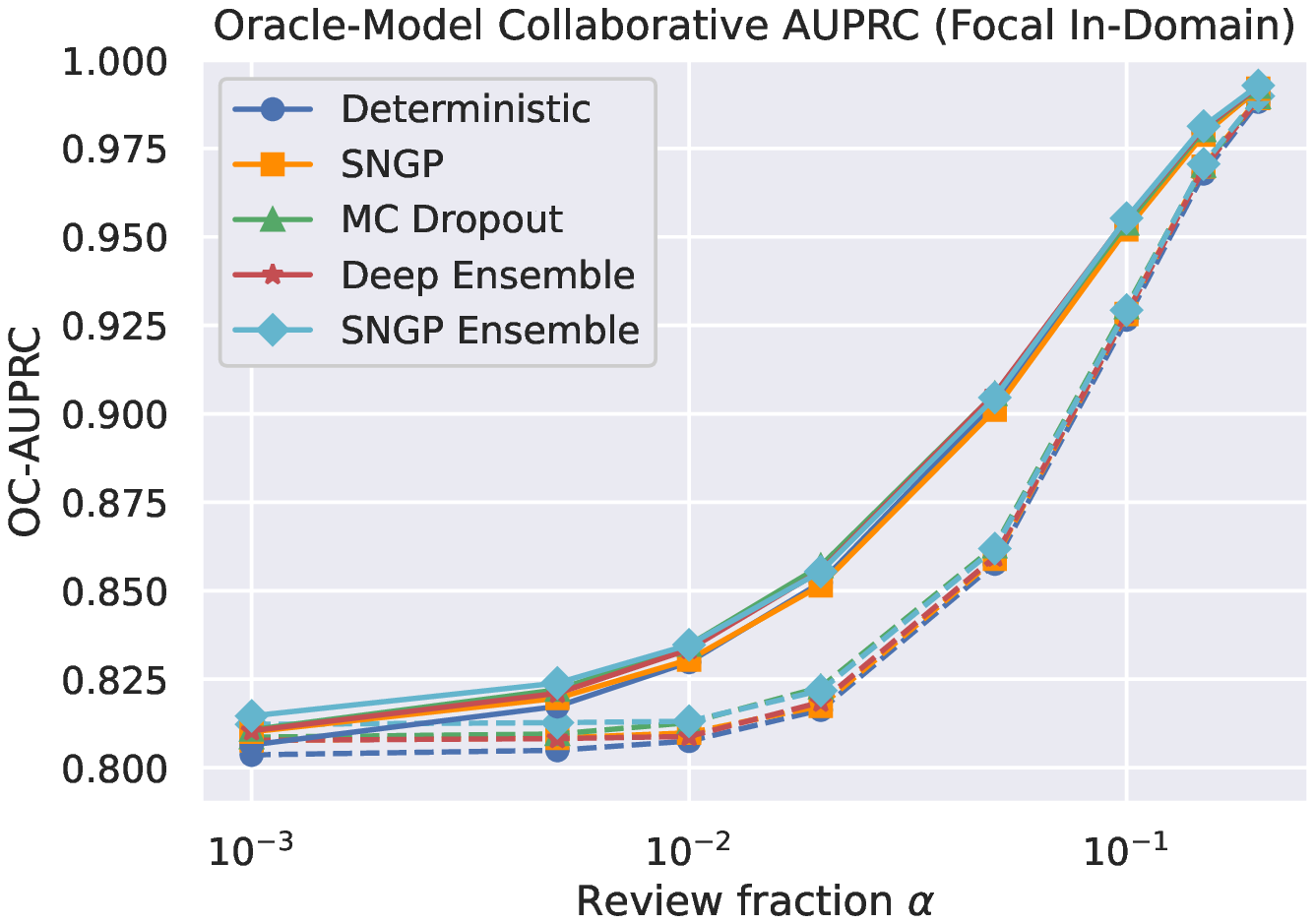}
\caption{Oracle-model collaborative AUPRC as a function of review fraction, trained with cross-entropy (left) or focal loss (right) and evaluated on Wikipedia Toxicity corpus (in-domain test environment). \textbf{Solid Line}: uncertainty-based strategy. \textbf{Dashed Line}: toxicity-based strategy. Overall, the SNGP Ensemble with focal loss using the uncertainty review performs best across all $\alpha$. Restricted to cross-entropy loss, the Deep Ensemble using uncertainty-based review performs best until $\alpha \approx 0.1$, when some of the toxicity-based reviews (e.g.\ SNGP Ensemble) begin to outperform it.
\label{fig:ocauprc_ind_focal}
}
\end{figure*}

\begin{figure*}[ht]
\centering
\includegraphics[width=0.4\textwidth]{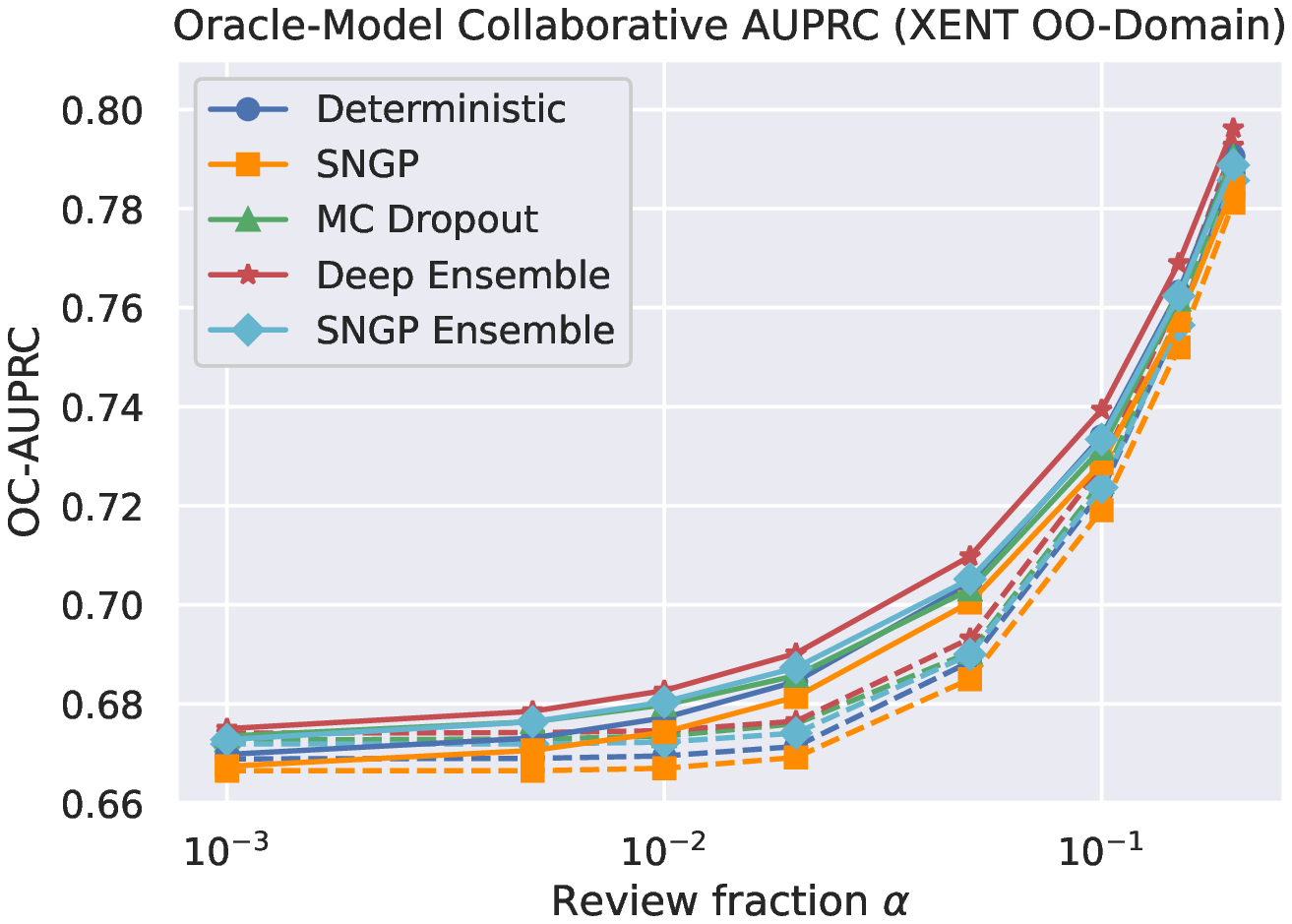}
\includegraphics[width=0.4\textwidth]{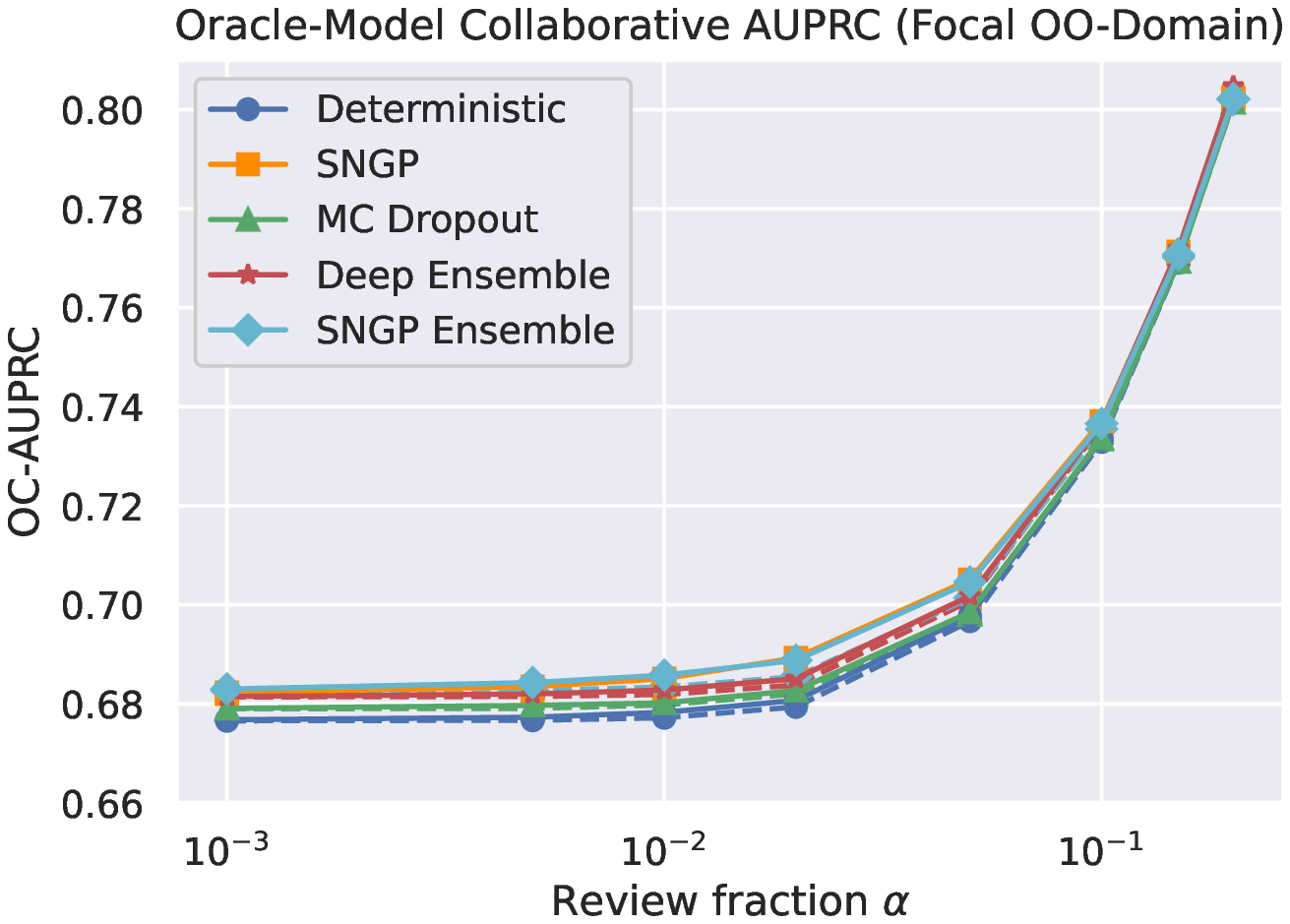}
\caption{Oracle-model collaborative AUPRC as a function of review fraction, trained with cross-entropy (left) or focal loss (right) and evaluated on CivilComments corpus (out-of-domain deployment environment). \textbf{Solid Line}: uncertainty-based strategy. \textbf{Dashed Line}: toxicity-based strategy. Similar to the out-of-domain OC-AUROC results in \fig{ocauroc_ood_focal}, of the models trained with cross-entropy loss the Deep Ensemble performs best. Training with focal loss yields a small baseline improvement, but surprisingly results in the SNGP Ensemble performing best. The uncertainty-based review strategy uniformly outperforms toxicity-based review, though the difference is small when training with focal loss.
\label{fig:ocauprc_ood_focal}
}
\end{figure*}

\begin{figure*}[ht]
\centering
\includegraphics[width=0.4\textwidth]{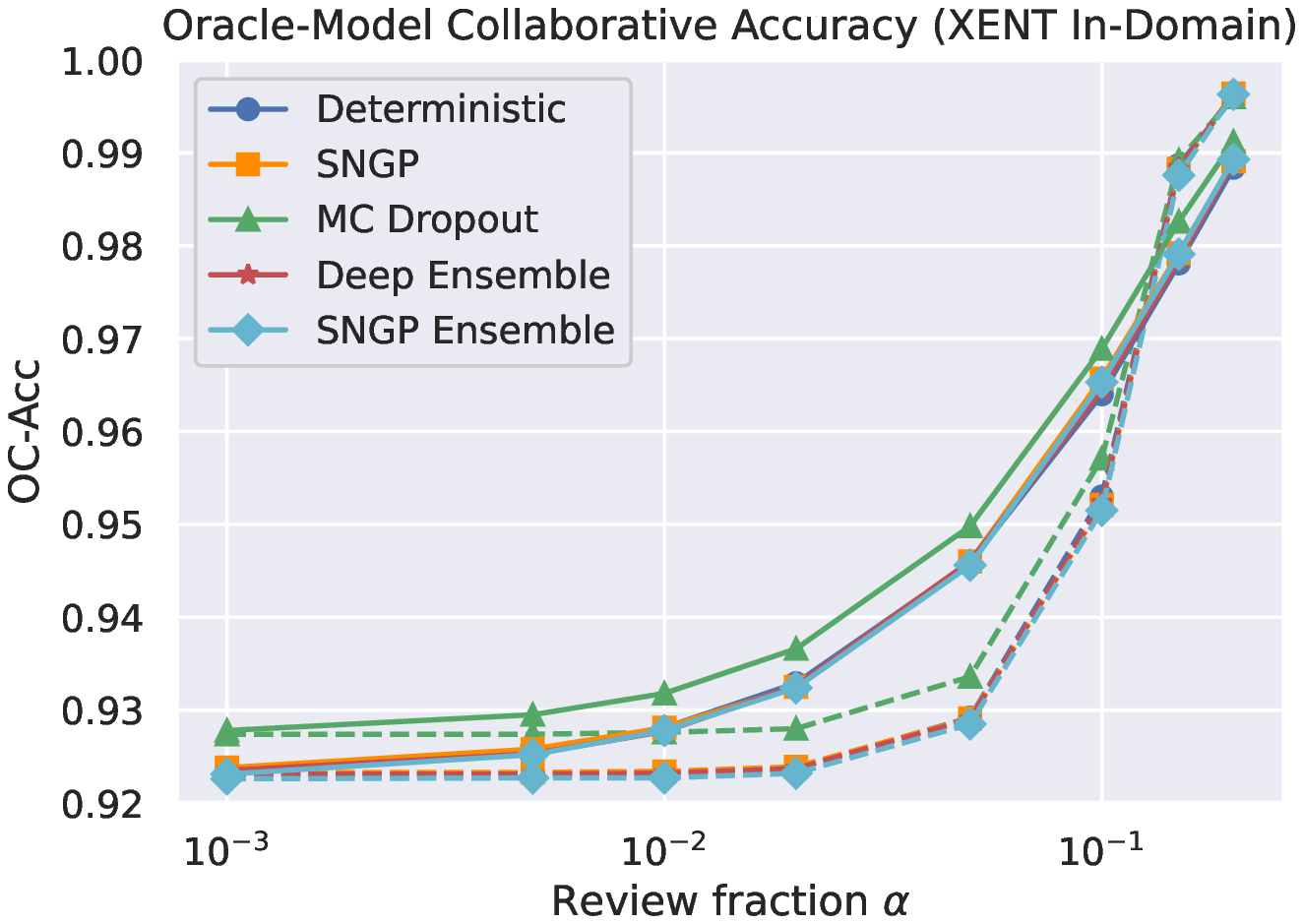}
\includegraphics[width=0.4\textwidth]{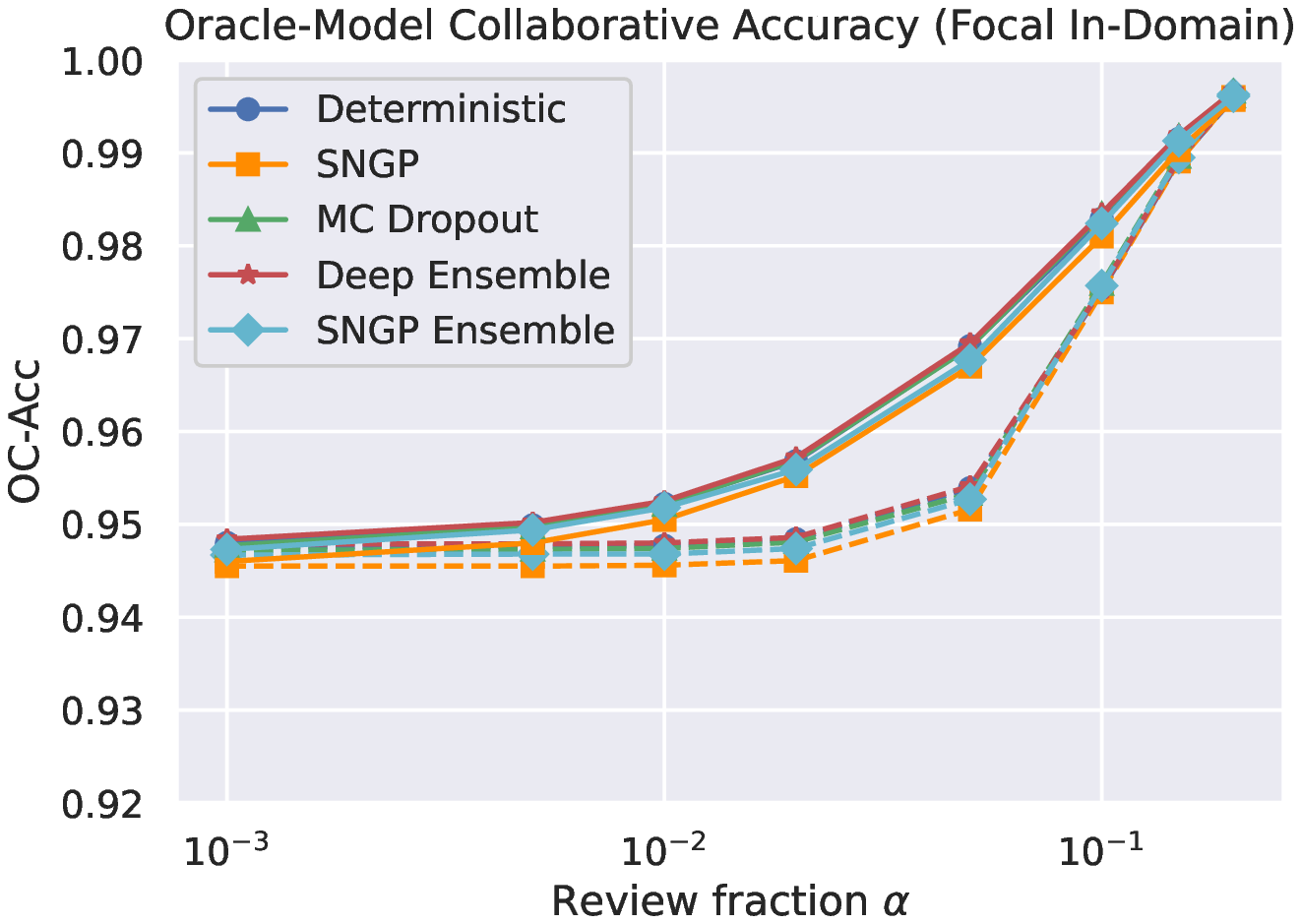}
\caption{Oracle-model collaborative accuracy as a function of review fraction, trained with cross-entropy (left) or focal loss (right) and evaluated on Wikipedia Toxicity corpus (in-domain test environment). \textbf{Solid Line}: uncertainty-based strategy. \textbf{Dashed Line}: toxicity-based strategy. Focal loss yields a significant improvement, equivalent to using a $10\%$ review fraction with cross-entropy. For most review fractions (below $\alpha=0.1$), MC Dropout using the uncertainty review strategy performs trained with cross-entropy, while overall the Deep Ensemble with focal loss (again using the uncertainty review) performs best. For large review fractions ($\alpha > 0.1$), the toxicity-based review in fact outperforms the uncertainty review.
\label{fig:oca_ind_focal}
}
\end{figure*}

\begin{figure*}[ht]
\centering
\includegraphics[width=0.4\textwidth]{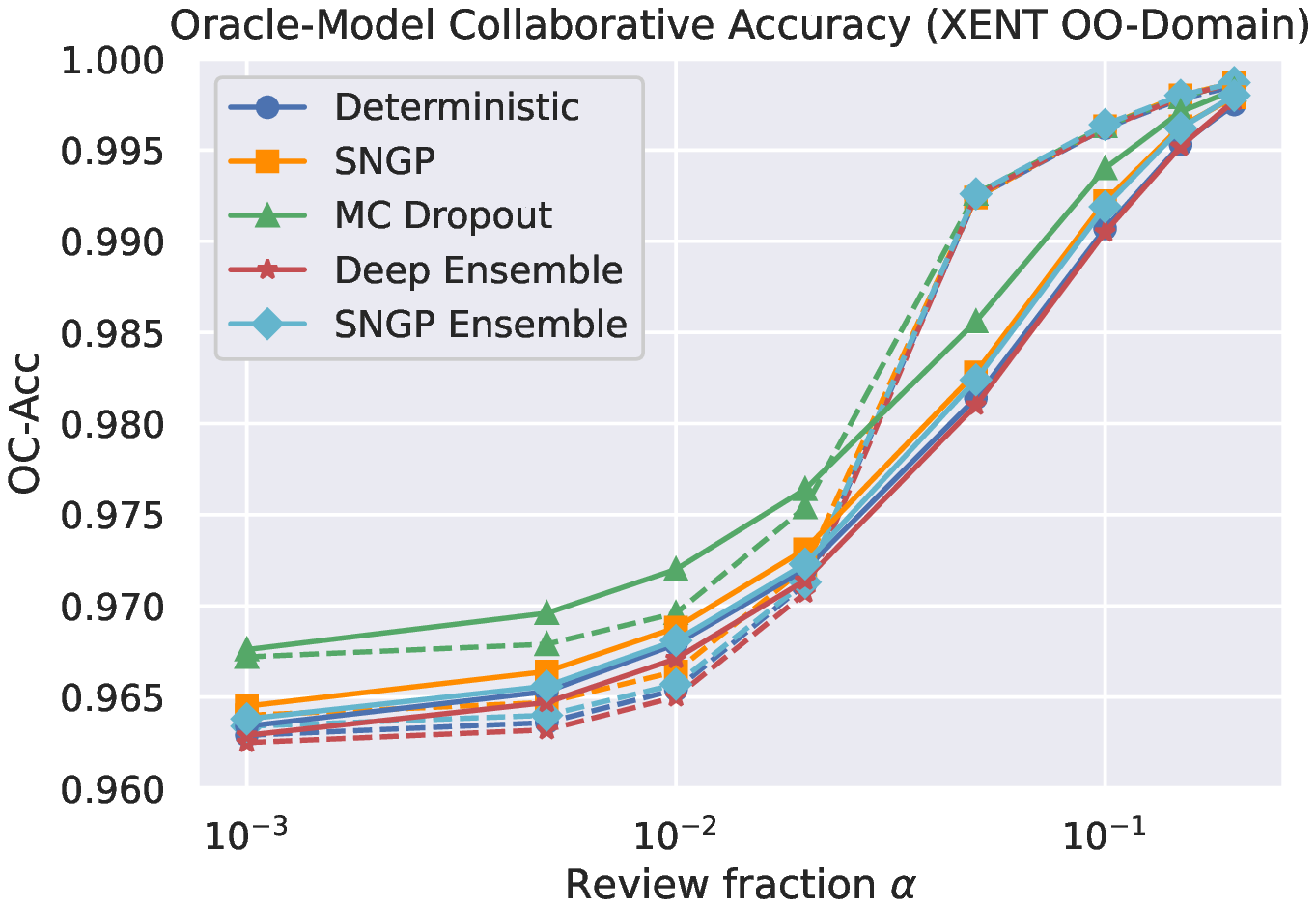}
\includegraphics[width=0.4\textwidth]{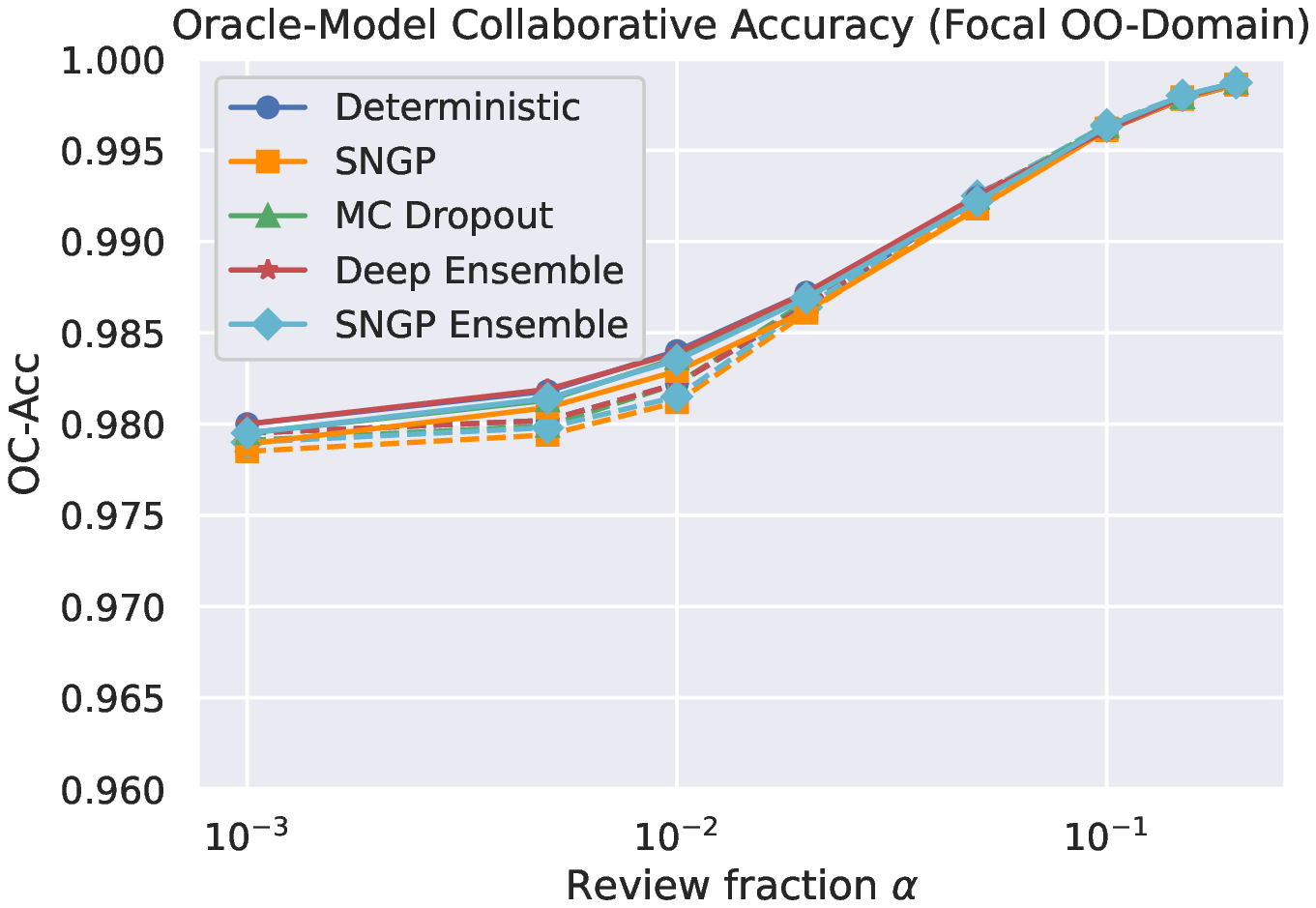}
\caption{Oracle-model collaborative accuracy as a function of review fraction, trained with cross-entropy (left) or focal loss (right) and evaluated on CivilComments corpus (out-of-domain deployment environment). \textbf{Solid Line}: uncertainty-based strategy. \textbf{Dashed Line}: toxicity-based strategy. Training with cross-entropy, MC Dropout using uncertainty-based review performs best until the SNGP Ensemble using the toxicity-based review overtakes it at $\alpha=0.05$. Training with focal loss gives significant baseline improvements (by mitigating the class imbalance problem); the Deep Ensemble is best for small $\alpha$ while the SNGP Ensemble is best for large $\alpha$. Despite these baseline improvements, they appear to come at a cost of collaborative accuracy in the intermediate region around $\alpha \approx 0.05$, where the SNGP Ensemble trained with cross-entropy briefly performs best overall, apart from that region the models with focal loss and the uncertainty-based review perform best (Deep Ensemble for $\alpha \le 0.02$, SNGP Ensemble for $\alpha \ge 0.1$).
\label{fig:oca_ood_focal}
}
\end{figure*}

\begin{figure*}[ht]
\centering
\includegraphics[width=0.4\textwidth]{plots/abstain_prec_ood_XENT_log.eps}
\includegraphics[width=0.4\textwidth]{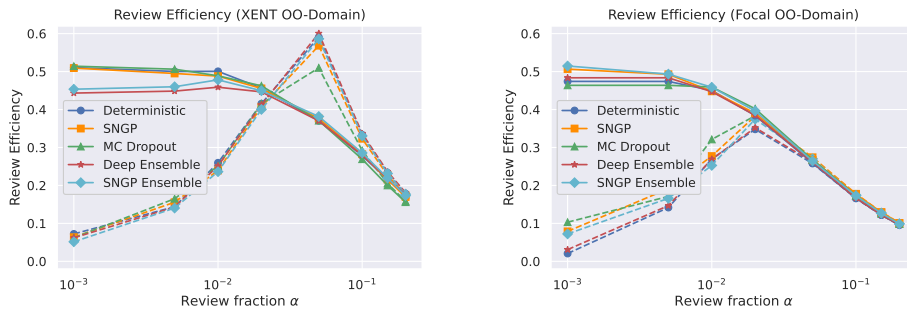}
\caption{Review efficiency as a function of review fraction, trained with cross-entropy (left) or focal loss (right) and evaluated on CivilComments corpus (out-of-domain deployment environment). \textbf{Solid Line}: uncertainty-based strategy. \textbf{Dashed Line}: toxicity-based strategy.
This is the plot for which we observe the largest crossover: training with cross-entropy, the efficiency for toxicity-based review spikes above the uncertainty-based review efficiency at $\alpha=0.02$ before converging back toward it with increasing $\alpha$. There is no corresponding crossover when training with focal loss; rather, the efficiencies of the two strategies converge at $\alpha=0.02$ instead.
\label{fig:abstain_prec_ood}
}
\end{figure*}

We give the results for the remaining collaborative metrics not included in the main paper in this appendix. These give a comprehensive summary of the collaborative performance of the models evaluated in the paper.
\tab{indomain_metrics_sep} and \tab{oodomain_metrics_sep} give values for all review fraction-independent metrics, both in- and out-of-domain, respectively. We did not include the ECE and calibration AUROC in the corresponding table in the main paper (\tab{indomain_metrics}) for simplicity.
Similarly, Figures \ref{fig:ocauprc_ind_focal} and \ref{fig:oca_ind_focal} show the in-domain results (the OC-AUPRC and OC-Acc), and the out-of-domain plots (in the same order, followed by Review Efficiency) are \fig{ocauprc_ood_focal}, \fig{oca_ood_focal}, and \fig{abstain_prec_ood}.

\begin{figure*}[ht]
\centering
\includegraphics[width=0.4\textwidth]{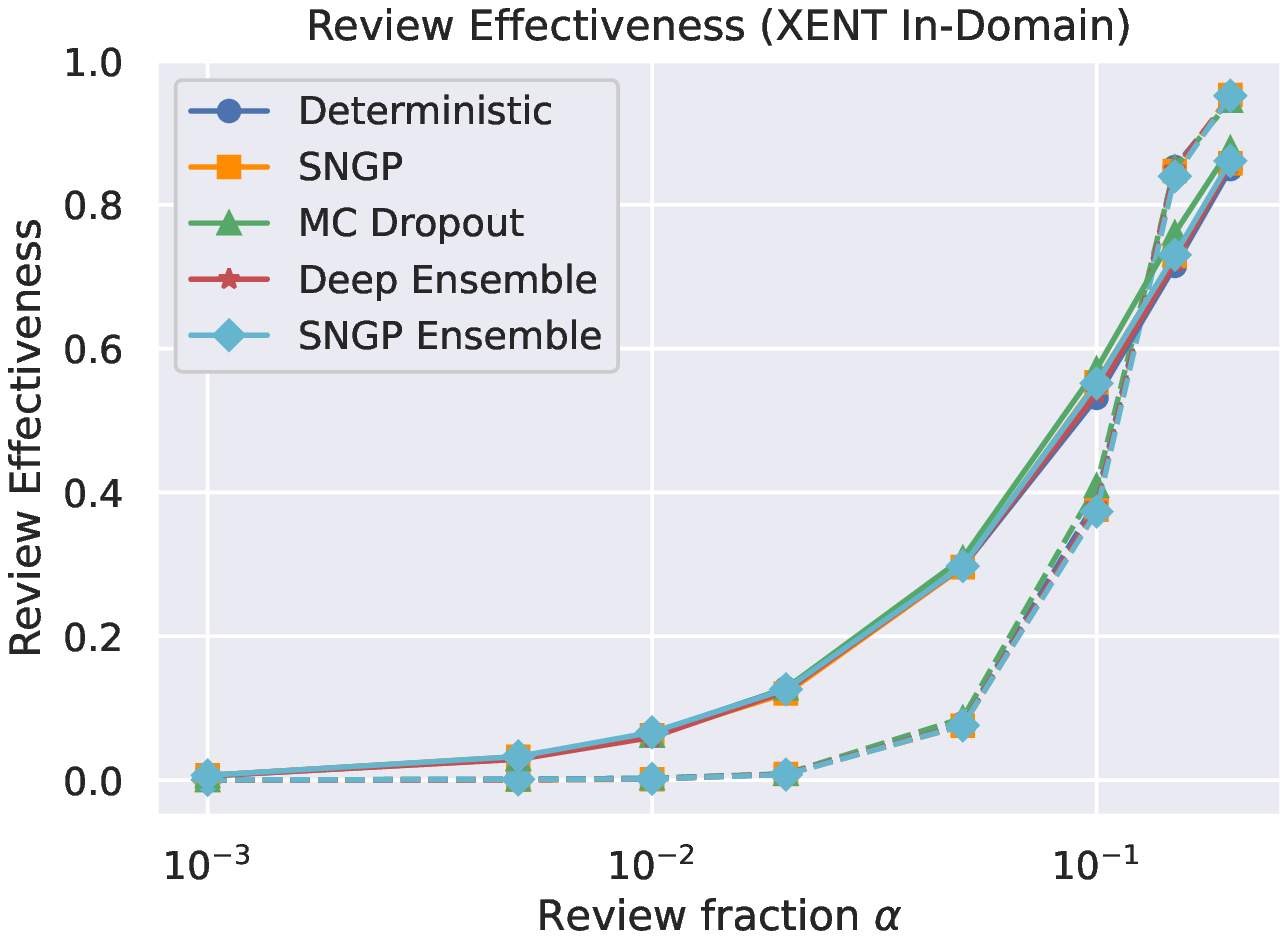}
\includegraphics[width=0.4\textwidth]{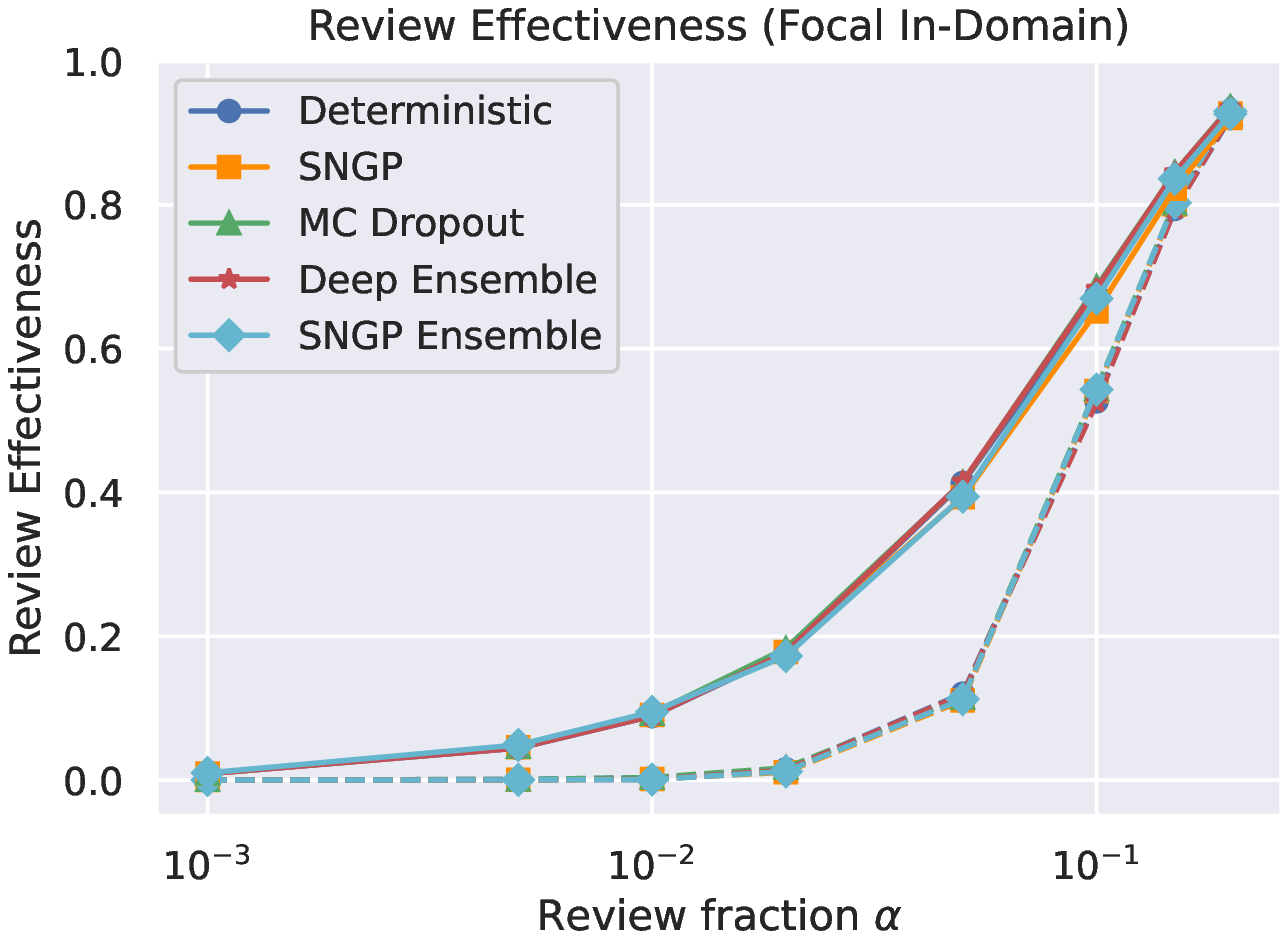}
\caption{Review effectiveness as a function of review fraction, trained with cross-entropy (left) or focal loss (right) and evaluated on Wikipedia Toxicity corpus (in-domain test environment). \textbf{Solid Line}: uncertainty-based strategy. \textbf{Dashed Line}: toxicity-based strategy. There is little difference between models here: the uncertainty-based review strategy successfully catches more incorrect model decisions until $\alpha \approx 0.15$.
\label{fig:abstain_recall_ind_focal}
}
\end{figure*}

\begin{figure*}[ht]
\centering
\includegraphics[width=0.4\textwidth]{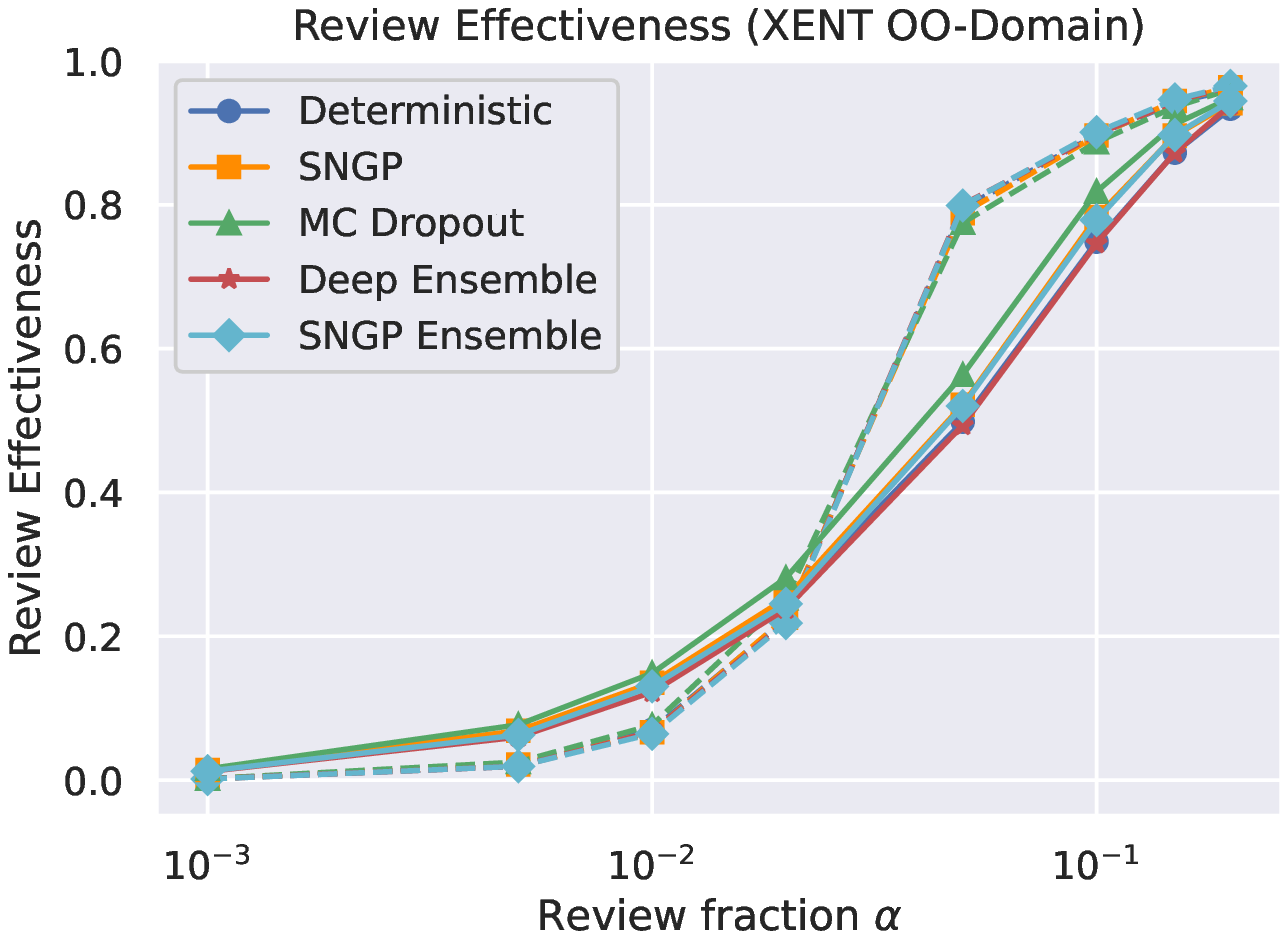}
\includegraphics[width=0.4\textwidth]{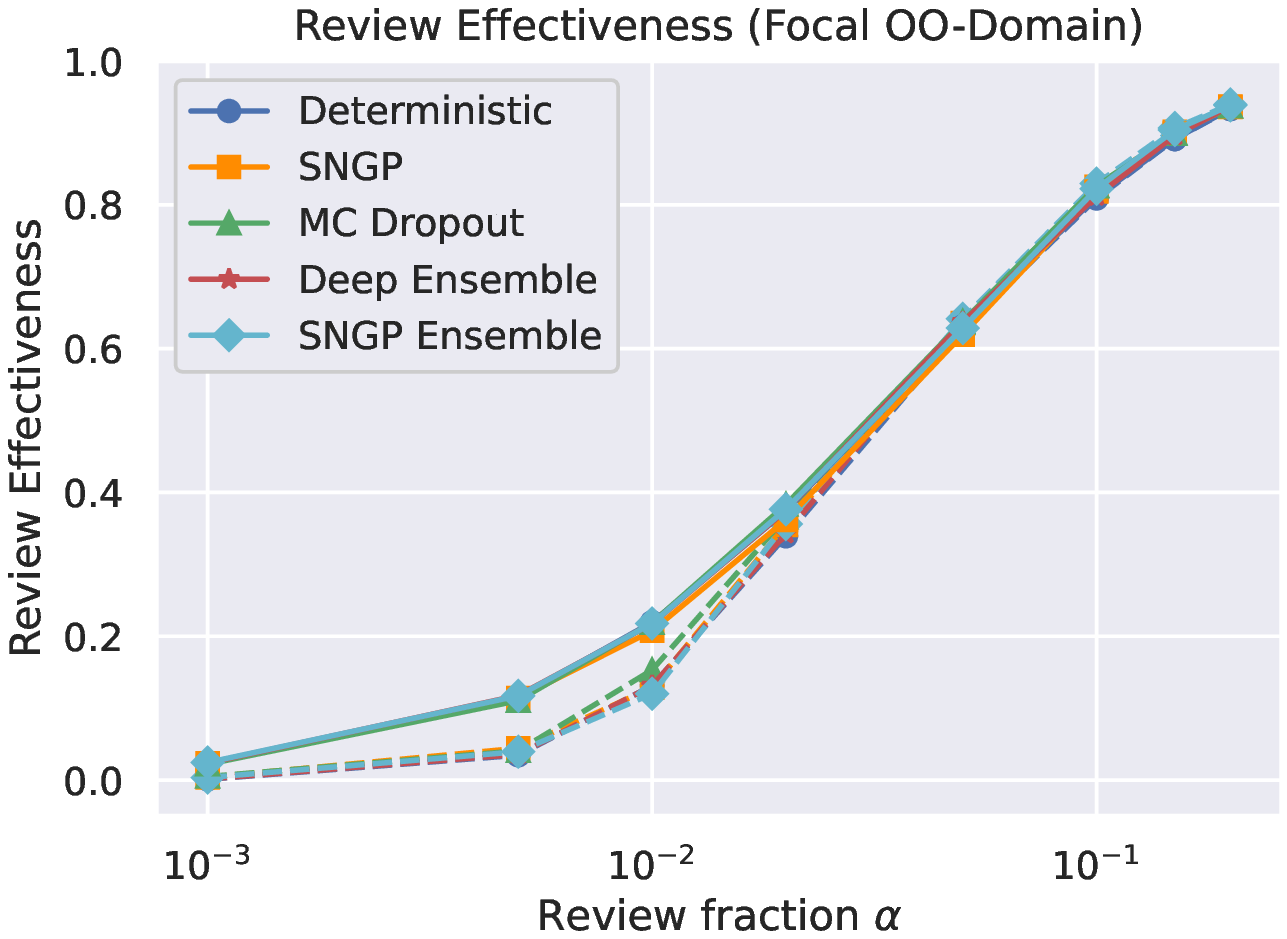}
\caption{Review effectiveness as a function of review fraction, trained with cross-entropy (left) or focal loss (right) and evaluated on CivilComments corpus (out-of-domain deployment environment). \textbf{Solid Line}: uncertainty-based strategy. \textbf{Dashed Line}: toxicity-based strategy. Here, the uncertainty review performs better until a crossover at $\alpha \approx 0.02$, much lower than in \fig{abstain_prec_ind_focal}. The SNGP Ensemble performs best with either cross-entropy or focal loss (slightly better with cross-entropy).
\label{fig:abstain_recall_ood_focal}
}
\end{figure*}

The in- and out-of-domain OC-AUROC figures are included in the main paper as \fig{ocauroc_ind_focal} and \fig{ocauroc_ood_focal}, respectively; the in-domain Review Efficiency is \fig{abstain_prec_ind_focal}. Additionally, we also report results on the Review Effectiveness metric (introduced in Section \ref{app:calib-auprc-discussion}) in Figures \ref{fig:abstain_recall_ind_focal}-\ref{fig:abstain_recall_ood_focal}. Similiar to Review Efficiency, we find little difference in performance between different uncertainty models, and that the uncertainty-based policy outperforms toxicity-based policy especially in the low review capacity setting.


\begin{figure*}[ht]
\centering
\includegraphics[width=0.4\textwidth]{plots/abstain_prec_ind_XENT_log.eps}
\includegraphics[width=0.4\textwidth]{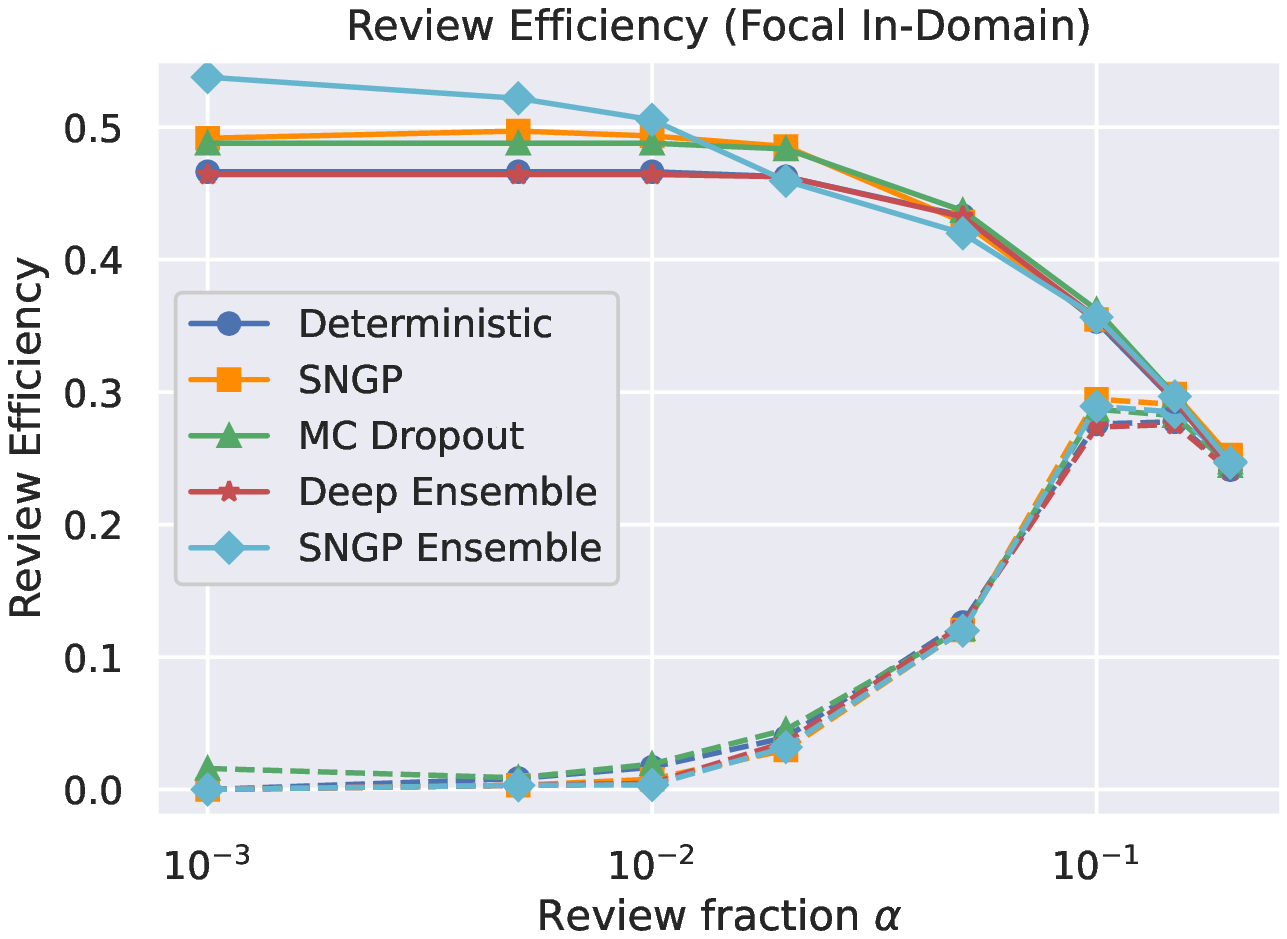}
\caption{Review efficiency as a function of review fraction, trained with cross-entropy (left) or focal loss (right) and evaluated on Wikipedia Toxicity corpus (in-domain test environment). \textbf{Solid Line}: uncertainty-based strategy. \textbf{Dashed Line}: toxicity-based strategy.
We see a small crossover in the in-domain performance when training with cross-entropy: the efficiency for toxicity-based review briefly exceeds the uncertainty-based review efficiency at $\alpha=0.02$ before converging back toward it with increasing $\alpha$. There is no corresponding crossover when training with focal loss.
\label{fig:abstain_prec_ind}
}
\end{figure*}

